\documentclass{article}

\usepackage{arxiv}

\usepackage[utf8]{inputenc}
\usepackage[T1]{fontenc} 
\usepackage{hyperref}     
\usepackage{url}          
\usepackage{booktabs}      
\usepackage{amsfonts}     
\usepackage{nicefrac}      
\usepackage{microtype}     
\usepackage{multirow}
\usepackage{graphicx}
\usepackage{lipsum}
\usepackage{float}
\usepackage{natbib}

\usepackage{tabularx,booktabs,ragged2e}
\usepackage{rotating}                      
\usepackage{newtxmath}                     
\usepackage[inline]{enumitem}              
\usepackage{caption,subcaption}            
\usepackage{eurosym}                       
\usepackage{xcolor}                        
\usepackage{url}                           
\usepackage{multirow}                      
\usepackage{float}                         
\usepackage{threeparttable} 
\usepackage[linesnumbered,boxed,lined]{algorithm2e} 
\usepackage{adjustbox}                              
\usepackage{bigdelim}
\usepackage{todonotes}
\usepackage{pdfpages}

\newcommand{\customfootnotetext}[2]{{
  \renewcommand{\thefootnote}{#1}
  \footnotetext[0]{#2}}}

\title{Fighting Sampling Bias: A Framework for Training and Evaluating Credit Scoring Models}

\author{
  Nikita Kozodoi\textsuperscript{*} \\
  Amazon Web Services \\
  Humboldt University of Berlin
  \And
  Stefan Lessmann\textsuperscript{$\dagger$} \\
  Humboldt University of Berlin \\
  Bucharest University of Economic Studies
  \And
  Morteza Alamgir\textsuperscript{$\ddagger$} \\
  Wolt Enterprises Deutschland GmbH
  \And
  Luis Moreira-Matias\textsuperscript{$\ddagger$} \\
  sennder technologies GmbH
  \And
  Konstantinos Papakonstantinou\textsuperscript{$\ddagger$} \\
  Kaizen Gaming
}

\begin{document}

\maketitle

\begin{abstract}
Scoring models support decision-making in financial institutions. Their estimation and evaluation are based on the data of previously accepted applicants with known repayment behavior. This creates sampling bias: the available labeled data offers a partial picture of the distribution of candidate borrowers, which the model is supposed to score.
The paper addresses the adverse effect of sampling bias on model training and evaluation.
To improve scorecard training, we propose bias-aware self-learning -- a reject inference framework that augments the biased training data by inferring labels for selected rejected applications.
For scorecard evaluation, we propose a Bayesian framework that extends standard accuracy measures to the biased setting and provides a reliable estimate of future scorecard performance.
Extensive experiments on synthetic and real-world data confirm the superiority of our propositions over various benchmarks in predictive performance and profitability. By sensitivity analysis, we also identify boundary conditions affecting their performance. 
Notably, we leverage real-world data from a randomized controlled trial to assess the novel methodologies on holdout data that represent the true borrower population.
Our findings confirm that reject inference is a difficult problem with modest potential to improve scorecard performance. Addressing sampling bias during scorecard evaluation is a much more promising route to improve scoring practices. For example, our results suggest a profit improvement of about eight percent, when using Bayesian evaluation to decide on acceptance rates.
\end{abstract}


\customfootnotetext{*}{This work does not relate to the author's position at Amazon}
\customfootnotetext{$\dagger$}{Corresponding author}
\customfootnotetext{$\ddagger$}{This work was conducted while the author was employed at Monedo Holding GmbH and does not relate to their current position.}

%
%

\section{Introduction}
\label{sec_introduction}

Assessing and mitigating the risk of money lending are crucial management tasks in the credit business. Machine learning (ML)-based scoring models support corresponding tasks by estimating credit default probabilities. Such estimates guide, for example, loan approval decisions, loss provisioning, and regulatory capital calculations. The paper studies the sampling bias problem, which routinely arises in the lending business and harms the quality of scorecard-based decisions.

To illustrate the problem, consider a loan applicant who seeks financing. The financial institution (FI) predicts the applicant's probability of repayment and decides the application accordingly. The FI can observe the repayment behavior of accepted clients and eventually determine if the debt was repaid. We refer to the groups of applicants who repay and default as \textit{good} and \textit{bad} risks, respectively. Sampling bias is a missing data problem. The FI cannot observe the repayment behavior of rejected clients \citep{wu_handling_2007}. 
More specifically, developing scorecards requires \textit{labeled} data with known outcomes (i.e., observed repayment status). Lacking labels for rejects, the FI uses data from previously accepted clients for scorecard development. This sample differs from the borrower population, which implies a bias \citep{banasik_sample_2003}. 

Credit scorecards impact borrowers, risk management, firm profitability, and, ultimately, society. This warrants concern about sampling bias. For example, in 2023, the total outstanding amount of consumer credit in the US exceeded \$5,019 billion \citep{federal2023}. Scorecards have played an important role in the approval of this amount. This shows how scorecards determine access to finance, which is a crucial factor for economic (in)equality \citep{wei_2016_credit}. 
Beyond retail, scorecards also inform corporate lending \citep{mues_ejor_transformer}, which emphasizes their prevalence and the potential implications of sampling bias. 

The objectives of this paper are to clarify the effect of sampling bias on scorecard predictions and lending operations and to develop methodologies that mitigate its adverse effect. In pursuit of this objective, the paper provides two methodological contributions. First, we introduce bias-aware self-learning (BASL), a reject inference (RI) framework that extends our previous work in \citet{kozodoi_2019_shallow} and addresses sampling bias during the training of a scoring model. Unlike existing data augmentation methods, we design BASL to prioritize prediction performance over debiasing and show this strategy to produce better predictions. 

A crucial step in the scorecard lifecycle concerns model evaluation. 
Quantitative estimates of scorecard accuracy also require labeled data, which again limits the data an FI can use for evaluation to the (biased) sample of accepted clients. Our second methodological contribution consists of a Bayesian evaluation framework that allows the calculation of arbitrary performance measures on a joint sample of accepted and rejected clients, which better reflects the borrower population. Drawing on prior knowledge, our framework overcomes the dependence on the unobservable labels of rejects and facilitates a reliable scorecard assessment under sampling bias. 

Beyond novel methodology, the paper contributes original insights into the severity of the sampling bias problem in credit scoring and the efficacy of bias mitigation strategies from extensive empirical experimentation using simulations and real-world micro-lending data. A unique feature of our real-world data is that it comprises an \textit{unbiased} sample of randomly approved loans. This sample represents the true operating conditions of a scorecard. Only seminal papers by \citet{banasik_sample_2003, banasik_credit_2005, banasik_reject_2007} had access to similar data, whereas the majority of RI studies employed the biased data from accepted applications to test their methodology. As we show in our empirical analysis, testing RI methods using a test set drawn from previously accepted clients is inappropriate and produces misleading results. Being in a position to overcome this problem highlights the empirical contribution of our study.

We proceed with discussing relevant background (Section~\ref{sec_background}) and related work (Section~\ref{sec_literature}) before presenting the novel methodologies for scorecard training (Section~\ref{sec_basl}) and evaluation (Section~\ref{sec_bayesian}). Next, we elaborate on our experimental design (Section~\ref{sec_setup}),  report empirical results (Section~\ref{sec_results}),  and discuss their implications (Section~\ref{sec_discussion}). Section~\ref{sec_conclusion} concludes the paper. The online appendix provides auxiliary results complementing the main paper.


\section{Theoretical Background and Problem Setting}
\label{sec_background}

This section formalizes sampling bias and the reject inference problem. Let $X \in \mathbb{R}^k$ denote a loan applicant. The matrix of the applicants' attributes is denoted as $\mathbf{X} = (X_1, ..., X_n)^\top$, and $\mathbf{y} = (y_1, ..., y_n)^\top$ is a random vector of binary labels, indicating if the applicant repays the loan ($y = 0$) or defaults ($y = 1$).
Suppose $\mathbf{X}$ and $\mathbf{y}$ have marginal distributions and a joint distribution denoted by, respectively, $\mathbf{P}_{X}$, $\mathbf{P}_{Y}$, and  $\mathbf{P}_{XY}$. 
Given a set of independent and identically distributed applications $D = \{(\mathbf{X}, \mathbf{y})\}$ with $(\mathbf{X}, \mathbf{y}) \sim \mathbf{P}_{XY}$, a financial institution uses a scorecard $f(X)$ that approximates $\mathbf{P}(y=1|X)$ to split $D$ into two subsets: accepts $D^{a}$ and rejects $D^{r}$, $D = D^a \sqcup D^r$. The repayment behavior is eventually observed for $D^a$. The labels of rejects remain unknown. Thus, $D$ exhibits missingness with respect to repayment outcomes $y$. Let $a \in \{0, 1\}$ denote a binary variable indicating if the applicant's repayment outcome is observed ($a = 1$) or missing ($a = 0$), which corresponds to whether the applicant was accepted.

The missing data literature distinguishes three types of missingness \citep{little2019statistical}. 
Labels are missing completely at random (MCAR) if $\mathbf{P}(a|X,y) = \mathbf{P}(a)$. 
Labels are missing at random (MAR) if $\mathbf{P}(a|X,y) = \mathbf{P}(a|X)$, implying that label missingness depends on the applicants' attributes but not on the repayment status. 
When labels are missing not at random (MNAR), the missingness also depends on $\mathbf{y}$ due to unobserved factors, which cannot be explained by the attributes. Formally, the data exhibits MNAR if $\mathbf{P}(a|X,y) \neq \mathbf{P}(a|X)$.

In credit scoring, we can ignore MCAR, which corresponds to a random acceptance policy. Approving loans using a scorecard causes $D^{a}$ to have different empirical distributions compared to $\mathbf{P}_{XY}$, $\mathbf{P}_{X}$ and $\mathbf{P}_{Y}$ and creates sampling bias. In practice, it is difficult to distinguish MNAR and MAR because the unobserved factors are not accessible. MAR occurs if a FI does not use external information apart from $\mathbf{X}$ to make acceptance decisions (e.g., always relies on predictions of the same scorecard). If scorecards and their underlying data change over time, we can still assume MAR if all changes have been documented and the policy by which loans were approved is completely known. 
On the other hand, cases like early repayment of loan obligations lead to censoring and, in turn, confounding, which implies MNAR. Another cause of MNAR is overwriting of scorecard recommendations based on attributes not included in $\mathbf{X}$. For example, applicants with a County Court Judgment may be manually rejected even if the scorecard prediction is positive \citep{banasik_reject_2007}. Such actions tie label missingness to factors unknown to the model $f(X)$. MNAR can also occur when some of the features in $\mathbf{X}$ included in a previous scorecard can no longer be used by a FI (e.g., due to new data privacy regulations or changes in data providers). 

Prior work provides several approaches for missing value imputation under MAR and MNAR to correct sampling bias \citep{little2019statistical}. However, full debiasing might not be the most suitable strategy in a credit scoring context. 
Facing a prediction problem, the primary goal of addressing sampling bias in the training data $D^a = \{(\mathbf{X}^a, \mathbf{y}^a)\}$ is to facilitate accurate approximation of $\mathbf{P}(y=1|X)$. Full debiasing of $D^a$ achieves this goal but may be difficult to achieve; especially under MNAR and/or when a large number of features complicate density estimation, which many imputation techniques entail \citep{Peng_ISR_22}. Full debiasing may also be dispensable in the pursuit of higher predictive accuracy. For example, under MAR, posterior probability models such as logistic regression do not require bias correction \citep{banasik_reject_2007}. More generally, given that the true missingness mechanism is unknown in practice, approaches for debiasing the training data can introduce noise. This diminishes the beneficial effect of reducing bias by introducing variance, leaving the net effect on the quality of the approximation of $\mathbf{P}(y=1|X)$, that is scorecard accuracy, unclear. We examine the relationship of sampling bias and scorecard accuracy in Appendix C.3 and C.4, and evidence a trade-off. A small reduction of bias can unlock sizeable gains in scorecard performance whereas over-emphasizing debiasing produces less accurate predictions. 

Apart from affecting scorecard training, sampling bias also impedes model evaluation under MAR and MNAR. A holdout set $H^a$ drawn from $D^a$ is not representative of $D \sim \mathbf{P}_{XY}$ because $D^a$ contains applications predicted as least risky. Thus, evaluating $f(X)$ using the data of accepts will overestimate scorecard performance \citep{banasik_sample_2003}.

Consolidating the theoretical considerations, we formalize our research goals as follows: given a set of labeled accepts $D^{a}$ and unlabeled rejects $D^{r}$, we aim to: (i) infer a function $f(X)$ that approximates $\mathbf{P}(y=1|X)$ and generalizes well over applications from $\mathbf{P}_{XY}$ and (ii) estimate the predictive performance of $f(X)$ over applications from $\mathbf{P}_{XY}$. 
Exploiting the information in $D^{r}$, we aim to address sampling bias in both tasks and improve, respectively, scorecard performance and the accuracy of predictions of scorecard performance. Facing uncertainty related to the type of missingness, we perform sensitivity analysis to test our propositions in MAR and MNAR settings.

\section{Related Work}
\label{sec_literature}

Sampling bias is extensively studied in different streams of literature with different perspectives. 
Examples include work on missing data problems due to non-random sampling \citep[e.g.,][]{Peng_ISR_22} or on policy learning \citep[e.g.,][]{athey_2021_econometrica}.
Further, a rich literature on sampling bias and RI in credit scoring exists \citep[e.g.,][]{Hand_ever_work_1993, banasik_reject_2007, Anderson_jors_montecarlo}.
Facing a large and diverse set of prior work, this section sketches related approaches in domains different from credit risk while also elaborating on relationships with prior RI studies and how we contribute to the field. 
We augment the discussion with a detailed tabular analysis of relevant papers in Appendix A of the online companion, where Table A.1 concentrates on sampling bias correction in general and Table A.2 focuses on RI studies in credit scoring.

%
%

\subsection{Training under Sampling Bias}
\label{sec_literature_training}

Representation change is a family of bias correction methods applied in the data prepossessing stage before training a corrected model. They assume MAR and use a mapping function $\Phi$ to project features into a new representational space $\mathbf{Z}$, $\Phi: \mathbf{X} \xrightarrow{} \mathbf{Z}$, such that the training data distribution over $\mathbf{Z}$ is less biased and $\Phi(X)$ retains as much information about $X$ as possible. A suitable representation is found by maximizing a distribution similarity measure such as the distance between the distribution moments in a kernel space \citep{borgwardt_integrating_2006}. 
Recently, generative models have gained popularity in learning feature transformations. For example, \citet{atan_deep-treat_2018} train a deep autoencoder with a mismatch penalty and extract the corrected data representation from the bottleneck layer. A potential disadvantage for credit scoring concerns interpretability. Regulatory compliance requires FIs to ensure comprehensible scoring models, which is difficult when employing transformed feature sets. 

Model-based bias correction methods adjust a learning algorithm to account for sampling bias. In his pioneering work, \citet{heckman_sample_1979} proposed a two-stage least-squares model for the MNAR setup. This model simultaneously estimates an outcome equation and a sample selection process, which facilitates eliminating bias in the estimated model parameters to yield consistent estimates. \citet{meng1985cost} developed a bivariate probit model with non-random sample selection for setups where the outcome variable is binary. Their model represents a sound approach for credit scoring under assumptions of MNAR and normally distributed residuals in the estimated equations. Notably, the work of \citet{heckman_sample_1979} and \citet{meng1985cost} exemplifies the focus in much prior work on missing data in that they emphasize unbiasedness in the estimates of model parameters. Many ML models are nonparametric and do not explicate internal parameters. Thus, our main interest is the impact of sampling bias on model predictions and their evaluation. We consider the bivariate probit models as a benchmark in our empirical analysis (see Section~\ref{sec_results_real}) and, as a byproduct of our research, suggest ways to examine bias in ML models analogously to how it is done in linear models using explainable AI tools (see Appendix D.2).

Another research stream considers mixture models for bias correction \citep[e.g.,][]{feelders_credit_2000}. These operate under the MAR assumption and treat the data as drawn from a mixture of two distributions: training and population. Learning from both samples, mixture models infer labels of new examples using the conditional expectation-maximization algorithm.  
A disadvantage of model-based methods is that they are embedded in a learning algorithm, which requires a specific classifier. Previous work has mostly focused on linear and parametric models with particular assumptions. Yet, there is evidence that other non-parametric approaches, such as XGB, perform better in credit scoring \citep[e.g.,][]{gunnarsson2021deep}.

Reweighting is another method that rebalances the training loss towards representative examples. Weights of the training examples, 
known as importance weights or propensity scores, can be computed as a ratio of two distribution densities: $w(X) = p_D(X) / p_{D^a}(X)$. High values of $w(X)$ indicate that $X$ is more likely drawn from $\mathbf{P}_{XY}$ and is, therefore, more important for training. Importance weights can be estimated in several ways \citep[e.g.,][]{huang_correcting_2006} and employed during training using, e.g., weighted least squares.
Since reweighting relies on attributes in $\mathbf{X}$, it assumes MAR and cannot correct for MNAR. For example, a reweighted training set still consists of previously accepted clients and misses certain distribution regions populated by rejects only. However, reweighting can still be helpful under MNAR as it may reduce errors in estimating a model from the training sample \citep{banasik_reject_2007}. 

The credit scoring literature has also explored the idea of data augmentation -- expanding the training sample by labeling and appending examples from $D^r$. The augmented sample covers a wider distribution region, which may reduce sampling bias. Prior work suggests different approaches that use a model trained over $D^a$ to label rejects. A classic example is hard cutoff augmentation (HCA), which labels rejects by comparing their predicted scores with the accepts-based model to a predefined threshold. Under sampling bias, reliance on the accepts-based model may increase the risk of error propagation when labeling rejects. Extrapolating predictions of the accepts-based scorecard on rejects is a valid technique for posterior probability classifiers under MAR but suffers from the omitted variable bias under MNAR \citep{banasik_sample_2003}. 

Parceling aims to improve upon HCA by considering rejects as riskier than accepts. Parceling splits rejects into segments based on the predicted score range, and labels reject within each range proportional to the assumed PD in that range. A decision-maker can then alter the probabilities compared to the ones observed within the same score range on $D^a$. Thus, parceling can work under MNAR if the decision-maker can correctly specify the change in the default probabilities across the considered groups of applicants.

This paper adds to the above literature by introducing BASL -- an RI framework that builds on self-learning-based data augmentation and incorporates extensions to account for sampling bias. The framework is model-agnostic and includes distinct regimes for labeling rejects and training a scorecard. This allows us to reduce the risk of error propagation during labeling rejects and employ a classifier with high discriminative power for screening new applications.

%
%

\subsection{Evaluation under Sampling Bias}
\label{sec_literature_evaluation}

Sampling bias also affects model evaluation. An estimate of model performance derived from a biased sample may not generalize to unseen data, which impedes standard model evaluation strategies such as cross-validation \citep{sugiyama_covariate_2007}. To address this problem, prior work proposes generalization error measures, whose asymptotic unbiasedness is maintained under sampling bias. This includes the modified Akaike information criterion \citep[MAIC,][]{shimodaira_improving_2000} and the generalization error measure suggested by \citet{sugiyama_input-dependent_2006}. Both measures rely on density ratio estimation 
, which may be difficult in higher dimensions. Measures like the MAIC are also limited to parametric learners.

Evaluation under sampling bias is also studied in off-policy evaluation research, which focuses on the evaluation of a policy (e.g., a classifier) in a contextual bandit setting with incomplete historical data \citep{athey_2021_econometrica}. In this setup, a policy reward depends on the action of a decision-maker and is only partially observed. A prominent policy evaluation method is importance-weighted validation, which reweights the reward towards more representative examples in the evaluation set using importance weights \citep{sugiyama_covariate_2007}. 

Reweighting produces biased estimates if the past policy is modeled incorrectly \citep{dudik2014doubly}. In our context, this implies that the attributes $\mathbf{X}$ do not explain prior acceptance decisions accurately, and the data exhibits MNAR. \citet{dudik2014doubly} recommend doubly robust (DR) estimators, which combine estimating importance weights with predicting policy reward (i.e., classifier loss). DR produces unbiased estimates if at least one of the modeled equations is correct. However, using DR in credit scoring is difficult. The contextual bandit setting considers a set of actions and assumes we observe a reward for the chosen action. DR can then impute the reward for other actions. In credit scoring, however, we do not observe a reward for rejected clients, which complicates the imputation of reward substantially. Also, measuring reward as classifier loss limits DR to performance measures calculated on the level of an individual loan. This prohibits using DR with popular rank-based metrics such as the area under the ROC curve (AUC). 

This paper introduces a Bayesian evaluation framework that remedies the adverse impact of sampling bias on model evaluation and provides a more reliable estimate of model performance. The framework is metric-agnostic and allows evaluating any scoring model on a data sample with labeled accepts and unlabeled rejects. The framework leverages prior knowledge of the label distribution among rejects and uses Monte-Carlo sampling to optimize calculations.

%
%

\subsection{Applications in Credit Scoring}
\label{sec_literature_applications}

Sampling bias has received much attention in credit scoring. Prior work focuses on scorecard training and tests debiasing methods including the Heckman model \citep[e.g.,][]{banasik_sample_2003}, data augmentation techniques such as HCA \citep[e.g.,][]{crook_does_2004}, and mixture models \citep[e.g.,][]{feelders_credit_2000}. A commonly used reweighting approach is banded weights, a cluster-based method that uses the bands of predicted probabilities of default to form clusters \citep[][]{banasik_credit_2005}. Augmentation by semi-supervised learning methods has also gained popularity \citep[e.g.,][]{li_reject_2017}. 
Table A.2 in Appendix A provides a detailed overview of empirical studies on RI.

The problem of evaluation under sampling bias has received less attention in the credit risk literature. 
Using a proprietary data set with labeled rejected applicants, \citet{banasik_sample_2003} illustrate the discrepancies between the accuracy estimates based on a sample from accepts and a representative holdout sample, and use the latter to judge the value of RI. To our knowledge, prior work has not considered techniques that mitigate the impact of sampling bias on model evaluation in the absence of such a sample.

A common characteristic of RI studies is that they use low dimensional data (see Table A.2). 
Recent studies emphasize the merit of alternative data such as applicants' digital footprints \citep[e.g.][]{Djeundje_2021_alternative_data} or the merit of crafting features from borrower networks \citep{maria_omega_21}. Further, few RI studies express the gains from RI in terms of profitability or use a proper representative sample to assess the benefits of bias correction; exceptions include \citep{banasik_credit_2005, chen_economic_2001}. We strive to address these limitations by employing a high-dimensional FinTech data set, evaluating performance on a representative sample from the borrowers' population, and examining the business impact of RI.

\section{Bayesian Evaluation Framework}
\label{sec_bayesian}

%
%

Estimating scorecard performance is crucial. 
Important policy changes such as purchasing additional data about applicants or changing the acceptance rate entail comparisons of scorecards' performance. Evaluating a scorecard on a biased validation set of accepted applicants produces biased performance estimates. The true scorecard performance in production will not match the expectations raised during evaluation, which may lead to an overestimation of its business value and suboptimal policy decisions. This section introduces the Bayesian evaluation framework that aims at mitigating the adverse effect of sampling bias on model evaluation.

Formally, the performance prediction task can be described as follows. Given a population of loan applicants $D = D^a \cup D^r$, 
we aim to evaluate the scoring model $f(X)$ on a representative holdout set denoted as $H, H \subset D$ to assess its true ability to infer the creditworthiness of new applicants.
Standard performance measures require knowledge of labels for all cases in $H$. Given that, in practice, only labels in $H^a = H \cap D^a$ are known, whereas labels of rejects in $H^r = H \cap D^r$ are missing, it is common practice to assess $f(X)$ on $H^a$ (i.e., evaluate based on accepts). Empirical results in Section \ref{sec_results} illustrate the inadequacy of this approach and motivate our proposition.

%
%

Specifically, our Bayesian framework extends standard performance metrics by adding rejects to the evaluation sample. Assume $f(X)$ is evaluated on $H$ using an arbitrary metric $\mbox{M}(f, H, \tau)$, where $\tau$ is a vector of metric meta-parameters (e.g., PD cutoff). Algorithm \ref{alg_bayesian} computes the Bayesian extension of $\mbox{M}$ denoted as $\mbox{BM}(f, H, \tau)$. Since the labels in $H^r$ are unknown, we use a prior of the label distribution among rejects $\mathbf{P}(\mathbf{y}^r, \mathbf{X}^r)$ to assign random pseudo-labels. This allows us to evaluate $f(X)$ on a representative sample consisting of labeled accepts and pseudo-labeled rejects.

\IncMargin{1em}
\begin{algorithm}[h]
    \small
    \SetAlgoLined
\SetKwInOut{Input}{input}\SetKwInOut{Output}{output}\SetKwInOut{Return}{return}
\Input{model $f(X)$, evaluation set $H$ with labeled accepts $H^a = \{(\mathbf{X}^a, \mathbf{y}^a)\}$ and unlabeled rejects $H^r = \{\mathbf{X}^r\}$, prior $\mathbf{P}(\mathbf{y}^r|\mathbf{X}^r)$, evaluation metric $\mbox{M}(f, H, \tau)$, meta-parameters $j_{max}, \varepsilon$}
\Output{Bayesian evaluation metric $\mbox{BM}(f, H, \tau)$}

$j = 0$; $\Delta = \varepsilon$; $E = \{\}$ \tcp*[r]{initialization}

\While{$(j \le j_{max})$ and $(\Delta \geq \varepsilon)$}{

    $j = j + 1$
    
    $\mathbf{y}^r = \mbox{binomial}(1, \mathbf{P}(\mathbf{y}^r|\mathbf{X}^r))$ \tcp*[r]{generate labels of rejects} 
    
    $H_j = \{(\mathbf{X}^a, \mathbf{y}^a)\} \cup \{(\mathbf{X}^r, \mathbf{y}^r)\}$ \tcp*[r]{construct evaluation sample}  
    
    $E_j = \frac{1}{j} \sum_{i=1}^{j} \mbox{M}(f, H_i, \tau)$ \tcp*[r]{evaluate f(X)}  
    $\Delta = E_j - E_{j-1}$  \tcp*[r]{check metric convergence}  
}

\Return{$\mbox{BM}(f, H, \tau) = E_j$} 
    \caption{Bayesian Evaluation Framework}
    \label{alg_bayesian}
\end{algorithm}
\DecMargin{1em}

The Bayesian framework employs Monte Carlo sampling. Each unknown label is drawn from a binomial distribution with the probability set to the prior for that rejected example. The Bayesian extension of the metric is then computed by averaging the metric values across multiple label realizations. The sampling iterations are terminated once the incremental change of the average value does not exceed a convergence threshold $\varepsilon$.

The accuracy of the estimated metric depends on the choice of the prior, which governs the sampling of class labels of rejects. We propose leveraging the attributes of examples in $D^r$ denoted by $\mathbf{X}^r$ and estimate the prior $\mathbf{P}(\mathbf{y}^r|\mathbf{X}^r)$ by rescoring rejects with a model that has been used to support loan approval decisions in the past or the original scores used in those decisions. 
The proposed framework stands on the trade-off of two components: benefits from evaluating a model on a more representative validation sample and additional noise in the labels of rejects based on an imperfect prior. As we establish through sensitivity analysis (see Section \ref{sec_results_synthetic}), gains from extending the evaluation sample will typically outweigh losses from the noise in the prior, which facilitates more accurate scorecard performance predictions.

It is useful to assess credit scorecards from different angles using different, complementary performance measures \citep{lessmann_benchmarking_2015}. In this regard, a notable feature of Bayesian evaluation is that it supports arbitrary performance measures. For example, characterizing the operating conditions of a scorecard, such as class priors and misclassification costs, loan characteristics (e.g., principal, interest rate), or estimates of the LGD and EAD can be incorporated in the parameter vector $\tau$ to approximate scorecards' profitability \citep{verbraken_development_2014}. 

\section{Bias-Aware Self-Learning Framework}
\label{sec_basl}

%
%

The BASL framework is based on a semi-supervised learning algorithm called self-learning \citep{levatic_self_2017}. 
Given a labeled set $D^a$ and an unlabeled set $D^r$, self-learning trains a supervised model over $D^a$. This model scores examples in $D^r$. Provided predictions exceed a specified threshold, examples are assigned the corresponding labels and appended to $D^a$. The model is then retrained on $D^a$ and used to score the remaining examples in $D^r$. This procedure repeats until a stopping criterion is met.

Sampling bias impedes the effectiveness of self-learning. First, labeling rejects based on the confidence of the accepts-based model may be misleading. Rejects come from a different distribution region, and the model can produce overconfident predictions, which become less reliable as the difference between the two samples increases. These errors are propagated through consecutive labeling iterations. Second, the accuracy of the assigned labels is at risk when using a strong learner, which may be prone to overfitting the biased training data. Third, using the same confidence thresholds for labeling \textit{good} and \textit{bad} rejects preserves the class ratio in the augmented labeled sample. In contrast, the \textit{bad} ratio in the population of applicants is higher than among accepts. Finally, employing commonly used stopping criteria based on the absence of examples with confident predictions may lead to exceeding the suitable number of labeling iterations, which risks overfitting and can strengthen the error propagation due to the bias.

%
%

\begin{figure}[h]
    \caption{Bias-Aware Self-Learning Framework}
    \label{fig_basl}
    \includegraphics[width = \textwidth , trim = {0 8.5cm 0 -0.1cm}, clip]{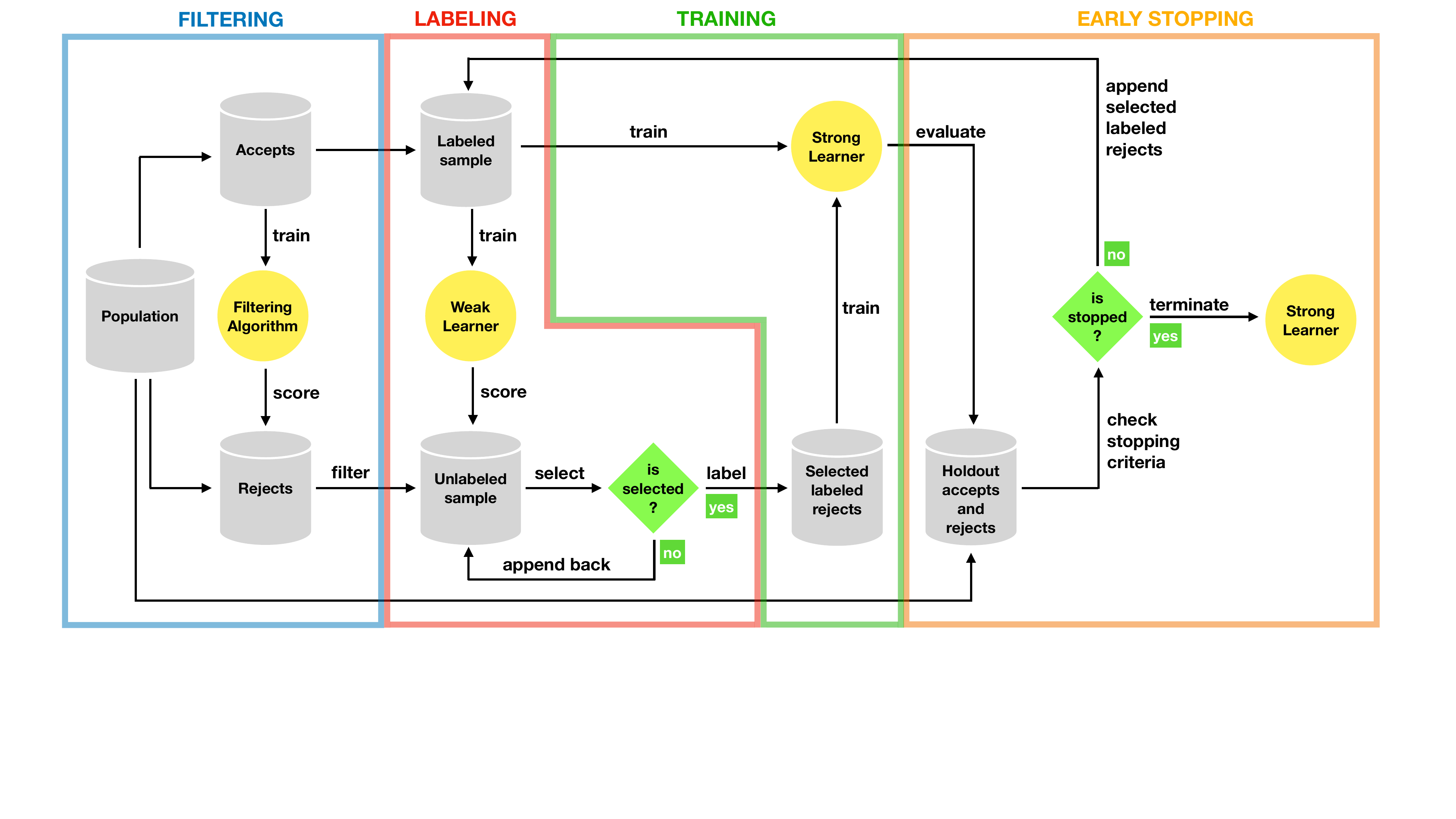}
        \footnotesize
        \textit{Notes:} Data splits into labeled accepts, unlabeled rejects, and a holdout set containing accepts and rejects. The filtering algorithm (isolation forest trained on accepts) selects rejects before labeling. Weak learner (LR) assigns labels to rejects. Strong learner (XGB) screens new loan applications. Selection criteria: (i) random sample of rejects with a rate $\rho$; (ii) Weak learner scores exceed confidence thresholds based on percentile thresholds $\gamma$ and imbalance multiplier $\theta$. Stopping criteria: (i) reaching maximum no. iterations $j_{max}$, (ii) performance of strong learner measured with the Bayesian framework is not improving; (iii) set of selected labeled rejects is empty.
\end{figure}

The BASL framework addresses the limitations of self-learning for RI with three extensions: (i) introducing a filtering stage before labeling, (ii) modifying the labeling and training regime, and (iii) introducing stopping criteria to handle sampling bias.
Figure \ref{fig_basl} summarizes the framework. Appendix B offers pseudo-codes for all BASL stages. Before elaborating on individual BASL stages, it is important to emphasize that 
BASL does not aim to eliminate bias in the training data. Full debiasing seems inappropriate because the repayment behavior of rejects from distant distribution regions is hard to estimate. 
Thus, we augment the training set with carefully chosen rejects for which the labeling model is confident and the data distribution is not too different from accepts. 
By doing so, we cannot expect the augmented sample to be free of bias. Rather, we aim to reduce bias while avoiding error propagation. The trade-off between bias reduction and scorecard accuracy is a crucial element of BASL, which we examine in detail in Appendix C.4 by constructing bias-accuracy Pareto frontiers with BASL variants that label different subsets of rejects.

\subsection{Filtering Stage}

This stage filters rejects in $D^r$ for two reasons. First, we aim to remove rejects from the most different part of the distribution compared to $D^a$. Removing such cases reduces the risk of error propagation since predictions of a model trained over accepts become less reliable when the distribution of rejects shifts further from the one observed during training. We also remove rejects that are most similar to accepts because labeling these would provide little new information and might even harm performance due to introducing noise. 

We estimate the similarity between rejects and accepts by scoring $D^r$ with a novelty detection algorithm trained over $D^a$. To remove the rejects most and least similar to accepts, we drop examples within the top $\beta_u$ and bottom $\beta_l$ percentiles of the predicted similarity scores. The threshold values $\beta = (\beta_u, \beta_l)$ act as meta-parameters of the filtering algorithm, which we implement using isolation forest, a scalable tree-based novelty detection algorithm suitable for high-dimensional feature spaces \citep{liu_isolation_2008}. \citet{xia_novel_2019} has also used isolation forest for filtering rejects. Interestingly, their approach differs fundamentally from BASL. They identify outliers among rejected clients and argue that these represent good applicants who were accidentally rejected. We conjecture that the fact that a rejected client differs in distribution from other rejects does not imply a repayment status and rather rely on distributional differences between accepts and rejects in BASL.


\subsection{Labeling Stage}

After filtering, we iteratively label selected rejects. We employ distinct regimes for labeling rejects and training the resulting scorecard and suggest scoring rejects using a learner with different inductive bias compared to that employed for scorecard construction. The labeling algorithm should provide well-calibrated predictions to select the appropriate confidence thresholds. Another desideratum of the labeling algorithm is that it should be less prone to overfitting the biased training sample. Using different algorithms for reject inference and scoring new applications also reduces the risk of amplifying the bias of the base classifier.

We use L1-regularized logistic regression (LR) as a weak learner for labeling rejects. The L1 penalty is introduced when working with high-dimensional data with noisy features. LR is a parametric learner that outputs probabilistic predictions. Appendix B.2 shows how LR predictions are better calibrated and take extreme values less frequently compared to a strong non-parametric learner such as XGB. Another benefit of LR over tree-based models such as XGB is its ability to extrapolate outside of the range of observed feature values, 
which is crucial since rejects stem from a different distribution region. Relying on recursive partitioning of the training data, tree-based methods cannot extrapolate outside feature value regions observed among accepts.

In each labeling iteration, we randomly sample $\rho m$ examples from the available set of $m$ rejects. Sampling aims at preventing overfitting by examining different regions of the distribution of rejects. Assuming that the currently deployed scorecard performs better than random, we expect the \textit{bad} rate in $D^r$ to be higher than in $D^a$. Therefore, we introduce the imbalance parameter $\theta$. We only label examples in the bottom $\gamma$ percentile and the top $\gamma \theta$ percentile of the distribution of scores predicted by the weak learner. This ensures that we select rejects with high confidence in the assigned labels and append more \textit{bad} examples than \textit{good} ones by setting $\theta > 1$. The latter helps to increase the \textit{bad} rate in the training sample to approximate the population distribution. The selected labeled rejects are removed from $D^r$ and appended to $D^a$. After the first iteration, we fix the absolute values of the confidence thresholds and use them in the following iterations.


\subsection{Training Stage}

At the end of each labeling iteration, we train a scoring model on the augmented labeled sample $D^a$ containing accepts and selected labeled rejects. The augmented sample covers a wider range of the feature space compared to the original sample of accepts. This reduces the effect of sampling bias on the trained model. The training stage benefits from using a strong base learner to develop a scorecard with high discriminative power to screen new applications. Following \citet{gunnarsson2021deep}, we use XGB as a base classifier for the resulting scorecard.


\subsection{Early Stopping}

The number of labeling iterations is controlled by the stopping criteria. We use our Bayesian framework to track the performance of the corrected scorecard across labeling iterations. At the end of each iteration, we evaluate the scorecard on a holdout sample containing labeled accepts and unlabeled rejects. Evaluating a model with the Bayesian framework is important as it allows us to account for the impact of sampling bias on evaluation. If the model performance does not improve, we stop labeling at this iteration and use the best-performing model as the final scorecard. We also specify the maximum number of labeling iterations $j_{max}$ and terminate BASL if there are no more rejects in $D^r$ for which predictions exceed the specified confidence thresholds. 

\section{Experimental Setup}
\label{sec_setup}

We perform a simulation experiment to illustrate sampling bias, demonstrate gains from our propositions, and investigate boundary conditions affecting their performance. Next, we test our methods on a high-dimensional microloan data set and estimate their business value. 

%
%

\subsection{Synthetic Data}
\label{sec_setup_synthetic}

We generate synthetic loan applications using two multivariate Gaussian mixtures:
\begin{equation}{
\begin{cases}
\mathbf{X}^g \sim  \sum_{c=1}^C \delta_c \mathcal{N}_k(\mathbf{\mu}_c^g, \mathbf{\Sigma}_c^g) \\
\mathbf{X}^b \sim  \sum_{c=1}^C \delta_c \mathcal{N}_k(\mathbf{\mu}_c^b, \mathbf{\Sigma}_c^b) \\
\end{cases}
}\end{equation}
where $\mathbf{X}^g$ and $\mathbf{X}^b$ are feature matrices of \textit{good} and \textit{bad} applications, and $\delta_c$, $\mathbf{\mu}_c$, and $\mathbf{\Sigma}_c$ are, respectively, the weight, mean vector and covariance matrix of the $c$-th mixture component. These parameters control the difference between the two applicant groups. 

To mimic the scorecard-based loan approval process, we devise a simulation framework called the acceptance loop hereafter. We assume a FI approves applications using a scorecard $f_a(X)$ that predicts 
$\mathbf{P}(y = 1|X)$. The FI accepts applicant $X$ if $f_a(X) \leq \tau$, where $\tau$ is a probability threshold. 
Suppose $D_j = \{(\mathbf{X}, \mathbf{y})\}$ is the batch $j$ of i.i.d. 
applications with
$(\mathbf{X}, \mathbf{y}) \sim \mathbf{P}_{XY}$ where 
$\mathbf{y}$ is unknown at the time of application. Acceptance decisions partition $D_j$ into 
$D_j^a = \{X_i \in \mathbf{X} | f_a(X_i) \leq \tau\}$ and 
$D_j^r = \{X_i \in \mathbf{X} | f_a(X_i) > \tau\}$ for accepts and rejects. 
Next, we reveal the labels in $D_j^a$ and retrain the scoring model on $D^{a} = \bigcup\limits_{j=1}^J D^a_j$, with $J$ denoting the total number of batches. We use the updated model to score new applications. Over time, $D^{a}$ grows in size with a bias towards accepts. 
We run the acceptance loop for $500$ iterations, generating a new batch of applications using the same distribution parameters each time.
We also draw a holdout sample from $\mathbf{P}_{XY}$ denoted as $H$. The sample $H$ facilitates assessing scorecard performance and bias correction methods on unseen data representative of the borrower population. Appendix C.1 further elaborates on the simulation.

Control over the data-generating process facilitates sensitivity analysis to clarify how the loss due to bias and gains from our propositions develop with changes in the environment. 
Following Section \ref{sec_background}, the sensitivity analysis comprises a gradual transition from MAR to MNAR. Other factors influencing the effectiveness of BASL include the strength of sampling bias, class imbalance, and the complexity of the classification task. Factors relevant for Bayesian evaluation include the validation sample distribution and the class prior for labeling rejects. The sensitivity analysis proposes measures for these factors and examines their impact on our propositions.

%
%

\subsection{Real Data}
\label{sec_setup_real}

The real data is provided by a FinTech company, Monedo, and consists of micro-loans issued to customers in Spain. The data includes 2,409 features characterizing loan applications. The binary target variable indicates whether the customer repaid the loan (\textit{good}) or experienced delinquency of at least three consecutive months (\textit{bad}). The data covers 59,593 applications, of which 39,579 were accepted, and 18,047 were rejected. In addition, we have access to a labeled, unbiased holdout sample of 1,967 applications that were approved at random. This sample includes applications the lender would normally reject and represents the through-the-door population of customers. The unbiased sample allows us to evaluate the performance gains from our propositions under the true operating conditions of Monedo. Table \ref{tab_dataset} summarizes the data and reveals that the \textit{bad} rate in the holdout sample is $1.7$ times higher than among accepts, which evidences sampling bias. Appendix D.1 provides additional information on the data including a description of the categories of features. Appendix D.2 offers an in-depth analysis of sampling bias in the real data and how it affects scorecard parameters, training, and evaluation. The corresponding results agree with those from synthetic data presented below in Section \ref{sec_results_synthetic}.

\begin{table}[h]
    \centering
    \small
    \caption{Real Data Summary}
    {\begin{tabular}{@{\extracolsep{15pt}} llll}
    \hline
     Characteristic & Accepts & Rejects & Holdout \\
    \hline
     Number of applications                     & 39,579 & 18,047  & 1,967 \\
    Number of features                             & 2,409  & 2,409   & 2,409 \\
     Percentage of \textit{bad} applications  & 39\%   & Unknown & 66\% \\
    \hline
\end{tabular}
    \label{tab_dataset}}
\end{table}

%
%

\subsection{Experiments}
\label{sec_setu_experiemtns}

We perform two experiments to test our propositions. 

Experiment I tests whether the Bayesian framework estimates the true scorecard performance more reliably than other evaluation strategies. To achieve this, we consider a performance prediction setup. We split accepts into training and validation sets and apply evaluation strategies to a scorecard trained on the training data. Each strategy aims to estimate the true scorecard performance on unbiased data from the borrower population. Ignoring sampling bias by evaluating on accepts is the baseline. DR and reweighting act as off-policy evaluation benchmarks.\footnote{Recall that differences between the off-policy evaluation setup and credit scoring prohibit the direct application of DR. Appendix F.2 details our implementation of an adjusted DR estimator that supports credit scoring.}
The Bayesian framework evaluates the scorecard on a merged validation set of accepts and unlabeled rejects. 
We assess the performance of an evaluation strategy by calculating the RMSE between the model performance estimates produced by that strategy over the experimental trials and the actual scorecard performance on the unbiased holdout sample.

Experiment II focuses on scorecard training. We correct the training set of accepts with one of the bias correction methods. We then train a scoring model over the corrected sample and evaluate it on a representative holdout sample. We compare BASL to several bias correction methods. Ignoring rejects serves as a baseline. Labeling rejects as \textit{bad} and bureau score-based labeling are simple augmentation techniques popular in credit scoring. HCA and parceling represent model-based augmentation methods. The Heckman model is another benchmark suited for MNAR and established in the credit scoring literature. We also implement reweighting with cluster-based weights. The bias-removing autoencoder is our representation change benchmark.

The simulation study allows us to conduct the experiments within the acceptance loop and aggregate the results over $100$ trials. Knowledge of the actual labels of synthetic rejects also allows us to implement an oracle model $f_o(X)$ trained on $D^{a} \cup D^{r}$. The oracle represents a scorecard that does not suffer from sampling bias and gives an upper performance bound. The real data does not support such a dynamic evaluation or an oracle solution. To improve the robustness of the results obtained on the real data, we aggregate performance over $100$ values from $4$ cross-validation folds times $25$ bootstrap samples of the holdout set. 
Further details on the data partitioning and meta-parameter values of bias correction methods are provided in Appendix E.

The simulation study and the real data experiment employ a set of complementary indicators to assess credit scorecards. The AUC and the Brier Score (BS) are widely used evaluation measures, which assess, respectively, the discriminatory ability of a scorecard and the degree to which its predictions are well-calibrated. Given that accepting a \textit{bad} applicant incurs higher costs than rejecting a \textit{good} applicant, we also consider the Partial AUC (PAUC). Summarizing the ROC curve on a limited range of cutoffs, the PAUC can account for asymmetric error costs. We compute the PAUC in the area of the ROC curve with a false negative rate in $[0, .2]$, which exemplifies a selective approval policy with low acceptance rates. Lastly, assuming FIs aim at minimizing losses while approving a specific percentage of applications, denoted as $\alpha$, we consider the \textit{bad} rate among accepts (ABR), where accepts are the top $\alpha \%$ applications with the lowest estimated PDs. Specifically, we integrate the ABR over acceptance between $20\%$ and $40\%$, which reflects historical policies at Monedo. For each metric, we use Algorithm \ref{alg_bayesian} to compute the Bayesian extension of that metric, which estimates performance over the unbiased holdout data $H$.

\section{Results}
\label{sec_results}

We begin with reporting empirical results from our main experiments and the sensitivity analysis when using synthetic data. Thereafter, we discuss the real data results and examine the business impact of our propositions.

%
%

\subsection{Synthetic Data}
\label{sec_results_synthetic}



\subsubsection{Simulation Results in a MAR Setup}
\label{sec_results_synthetic_basic}

\begin{figure}[h]
    \caption{Loss due to Sampling Bias and Gains from Our Propositions}
    \includegraphics[width = \textwidth, trim = {0 0 0 -0.1cm}, clip]{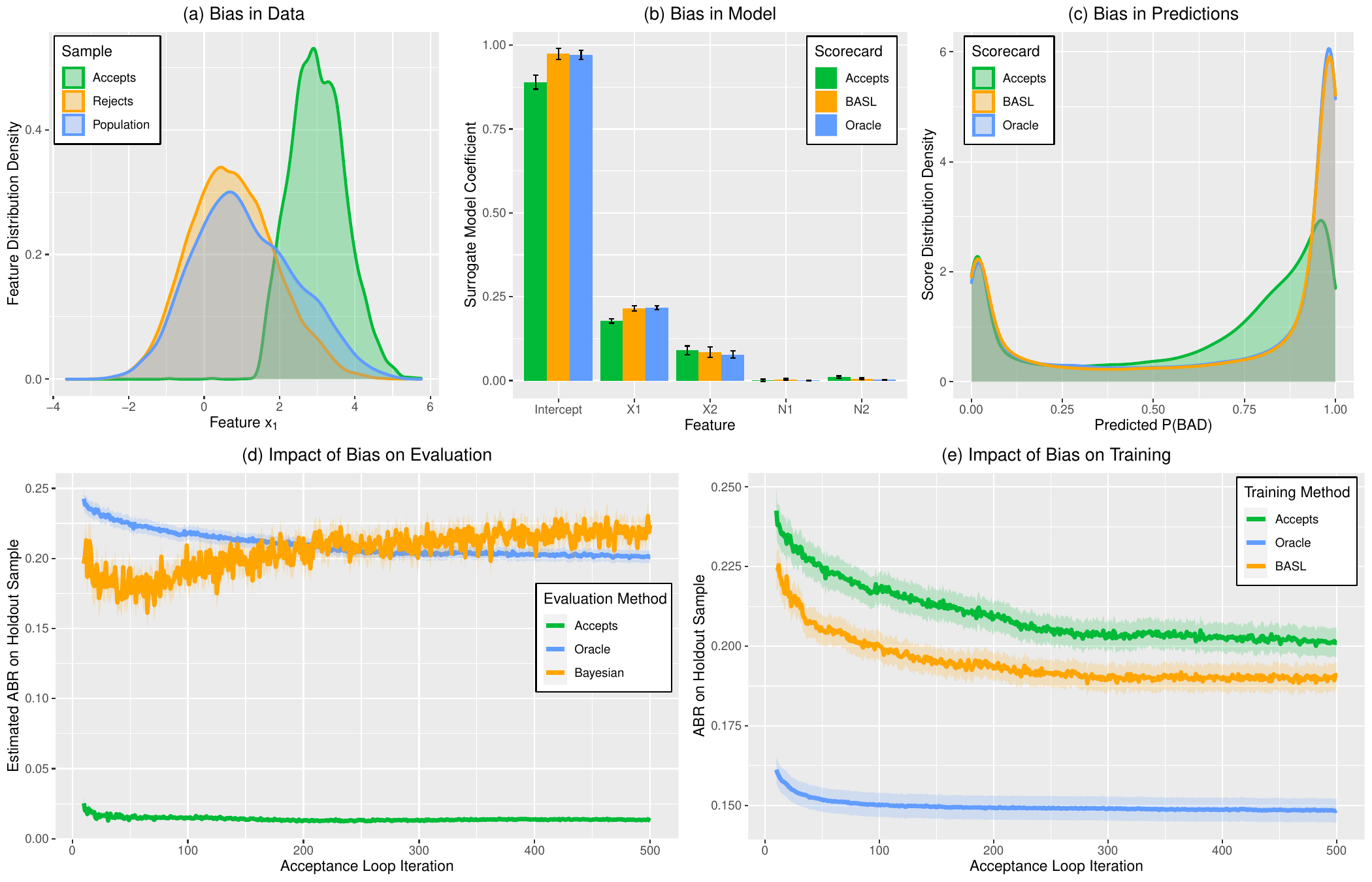}
    \label{fig_sim_results}
    \footnotesize \textit{Note:} The figure depicts sampling bias, its effect on scorecard training and evaluation, and gains from our propositions within the acceptance loop on synthetic data. Panel (a) shows bias in the distribution of one feature. Panel (b) visualizes bias in scorecards by comparing the coefficients of the corresponding linear surrogate models. Panel (c) shows bias in scorecard predictions for new loan applications in the holdout sample. Panels (d) and (e) depict the impact of bias on scorecard training and evaluation in terms of the ABR. 
\end{figure}

Figure \ref{fig_sim_results} illustrates sampling bias and its adverse effects on scorecard training and evaluation. Panel (a) compares the distribution densities of one of the synthetic features in $D^a$, $D^r$ and $H$. The results indicate differences in the distribution of $x_1$ in $D^a$ and $H$. The values of $x_1$ in $(-3, 1)$ are not observed among accepts in $D^a$, although the density peak in the unbiased set $H$ is located within this interval. This confirms that the training data of previously accepted clients are not representative of the population of loan applicants. 

Bias in the training data affects the scorecard behavior. We use a non-parametric XGB-based scorecard, which prohibits a direct inspection of the model parameters to illustrate the bias in the classifier. Regressing applicant features on the predictions of an XGB scorecard using linear regression, we obtain a surrogate model that approximates how XGB maps feature values to predictions. Panel (b) compares the coefficients of the surrogate models corresponding to the three XGB scorecards: (i) biased model $f_a(X)$ trained over $D^{a}$; (ii) oracle model $f_o(X)$ trained over $D^{a} \cup D^{r}$; (iii) model $f_c(X)$ corrected by labeling rejects with BASL. The results indicate that sampling bias affects the coefficients of surrogate scorecards and causes them to diverge from the oracle values. BASL partly recovers this difference, bringing the coefficients closer to the oracle. The bias in model parameters translates into a difference in the scores predicted by the scorecards. As illustrated in panel (c), $f_a$ provides more optimistic scores compared to $f_o$, whereas the distribution of scores produced by $f_c$ is more in line with that of the unbiased model.

Panel (d) of Figure \ref{fig_sim_results} depicts the results of Experiment I, which examines the impact of sampling bias on the scorecard evaluation. It compares the ABR of $f_a$ on the representative holdout sample $H$ (labeled as oracle performance) and the estimated ABR of $f_a$ using accepts-based vs. Bayesian evaluation. Evaluating $f_a$ on a sample from $D^a$ provides an overoptimistic estimate of around $.0140$. However, scoring new applications from $H$, we observe an ABR of around $.2095$. This gap illustrates that the accepts-based performance estimate is misleading due to sampling bias. The Bayesian framework provides a more reliable estimate of the ABR, reducing the mean RMSE between the actual and predicted ABR values from $.2010$ to $.0929$.

Panel (e) reports the results of Experiment II, depicting the effect of sampling bias on the scorecard performance. It compares the ABR of $f_a$, $f_o$ and $f_c$. The performance of $f_a$ gradually improves over acceptance loop iterations due to the increasing training sample size. The ABR of $f_o$ remains stable after 100 iterations and is consistently lower than that of $f_a$. The difference between $f_o$ and $f_a$ captures the loss due to sampling bias. The average ABR loss across 100 simulation trials is $.0598$. The model corrected with BASL consistently outperforms $f_a$ starting from the first iteration of the acceptance loop. Note that $f_c$ still suffers from the bias and does not reach the ABR of $f_o$. However, BASL consistently recovers about $25\%$ of the ABR loss compared to $f_a$.

We observe similar performance gains from Bayesian evaluation and BASL for other evaluation metrics (AUC, PAUC, BS) and report corresponding results for Experiment I and Experiment II in Appendix C.2 and C.3, respectively. 
Performing statistical hypothesis testing, we find that the observed gains are statistically significant. Friedman's rank sum tests reject the null hypothesis that our propositions perform the same as ignoring rejects in both experiments (p-values are lower than $2.2 \times 10^{-16}$). Pairwise Nemenyi post-hoc tests indicate that our propositions significantly outperform the accept-based training and evaluation at a 5\% significance level. 


\subsubsection{Sensitivity Analysis}
\label{sec_results_synthetic_sensitivity}

\begin{figure}
    \caption{Sensitivity Analysis: Bias-Aware Self-Learning}
    \includegraphics[width = \textwidth,  trim = {0 0 0 -0.1cm}, clip]{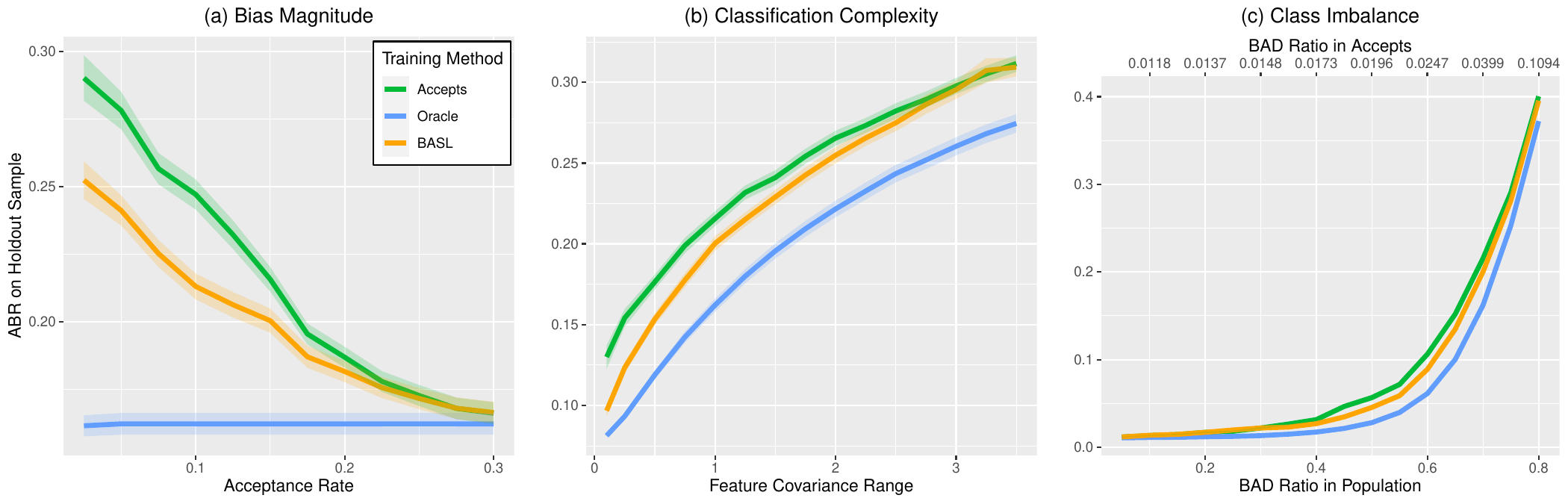}
    \footnotesize  \textit{Note: }The figure illustrates the impact of sampling bias on model training and benefits from RI with BASL as a function of three parameters: (i) bias magnitude controlled by the acceptance rate; (ii) classification task complexity controlled by the feature covariance range; (iii) class imbalance controlled by the \textit{bad} rate in the population.
    \label{fig_sensitivity_basl}
\end{figure}

This section examines factors that determine the effectiveness of our propositions to understand when their use is recommended and identify boundary conditions.
Figure \ref{fig_sensitivity_basl} depicts loss due to bias and gains from BASL across different simulations. Panel (a) examines the influence of the magnitude of sampling bias in the training data. The bias magnitude is controlled by the acceptance rate, which varies across financial products and markets. Lower acceptance results in a greater difference between $D^a$ and $D$. This raises the loss due to bias and the effectiveness of BASL to mitigate this loss. We find BASL to improve scorecards for acceptance rates below $20\%$, whereby this result originates from assuming a $30\%$ \textit{good} rate in the borrowers' population. Increasing the \textit{good} rate would lead to a slower decay of the loss due to bias and performance gains from BASL. Still, the analysis implies that BASL is more helpful for financial products with low acceptance rates, such as, e.g., installment loans for prime customers. The results also agree with \citet{crook_does_2004}, who find a negative relationship between the acceptance rate and performance gains from reweighting-based bias correction. 

Panel (b) studies the classification complexity and depicts the development of scorecard performance as a function of the feature covariance range. The elements of the feature covariance matrix are drawn randomly. A wider range of possible covariance values increases the classification complexity because loan applications of different classes tend to overlap more frequently in the feature space. The loss due to sampling bias is consistently present across the considered complexity range. Performance gains from BASL are higher in environments with a lower classification complexity and gradually diminish in more complex environments. This is because the pseudo-labels assigned to rejects are more accurate when class separation is easier. The ability to distinguish \textit{good} and \textit{bad} applicants is, therefore, an important factor governing the potential of RI. In practice, observed default rates can shed light on the complexity of the classification task associated with scoring applications for a financial product.

Panel (c) investigates the impact of class imbalance, which we control by the proportion of \textit{bad} applications in the population. The results suggest that any \textit{bad} rate in the population translates into class imbalance among accepts since the data is filtered by a scorecard. The loss due to bias shrinks when class imbalance becomes too strong. This comes from the ABR metric focusing on the least risky applicants, which are mostly \textit{good} due to high imbalance. BASL provides the largest gains at moderate imbalance between $2\%$ and $5\%$ among accepts. This imbalance level is sufficiently high so that an accepts-based model is not exposed to enough \textit{bad} risks but is not too severe to prohibit learning from the scarce number of \textit{bad} applications.

\begin{figure}[t]
    \caption{Sensitivity Analysis: Bayesian Evaluation}
    \includegraphics[width = \textwidth, trim = {0 0 0 -0.1cm}, clip]{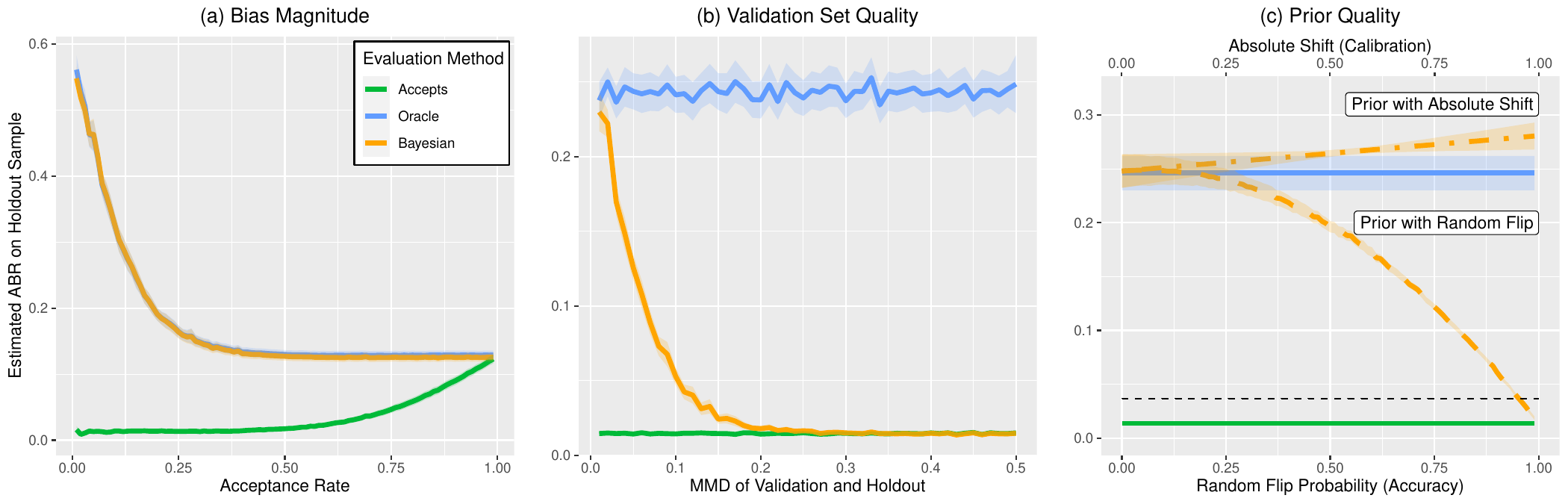}
    \label{fig_sensitivity_bayesian}
    \footnotesize \textit{Note: }The figure illustrates the impact of sampling bias on scorecard evaluation and benefits from the Bayesian framework as a function of three parameters: (i) bias magnitude controlled by the acceptance rate; (ii) validation set quality controlled by the accept/reject rate and measured by the MMD metric; (iii) prior quality controlled by injecting noise in the form of a random label flip or level shift. A dashed black line in panel (c) gives the ABR predicted by Bayesian evaluation when using the empirical \textit{bad} rate on accepts as a prior for labeling rejects.
\end{figure}

Turning to the Bayesian framework, panel (a) of Figure \ref{fig_sensitivity_bayesian} examines the effect of the acceptance rate on scorecard evaluation. To isolate this effect, we assume a perfect prior when calculating the Bayesian extension of the ABR. Under this assumption, the Bayesian framework estimates scorecard performance accurately across all acceptance rates. Similar to BASL, potential gains from Bayesian evaluation are higher at lower acceptance, as the inconsistency between the performance on accepts versus that on a representative sample becomes stronger. 

Calculating the Bayesian extension requires a validation sample of labeled accepts and unlabeled rejects. Panel (b) studies how the quality of this sample affects evaluation. We assess sample quality using the maximum mean discrepancy metric \citep[MMD,][]{borgwardt_integrating_2006}, which measures the similarity of the feature distribution in the validation set and the unbiased holdout set. The results reinforce accept-based evaluation to underestimate error rates substantially. To predict scorecard performance accurately, the Bayesian framework requires validation data that matches the target distribution in the holdout set. To ensure this, the validation sample should include accepts and rejects from the same period and match the accept/reject ratio in the holdout sample.

Panel (c) focuses on the class prior for labeling rejects. Reducing the quality of the estimate of the prior decreases the accuracy of Bayesian evaluation. The quality reduction can stem from diluting the accuracy or the calibration of the prior, which we simulate by injecting random flips or level shifts, respectively. The Bayesian framework provides more accurate performance estimates than accepts-based evaluation even under severely corrupted priors. This implies that using imperfect scores (e.g., from a currently used scorecard) as a prior for the labels of rejects still produces a better result than ignoring rejects during evaluation.


\subsubsection{Missingness Type}
\label{sec_results_synthetic_misingness}

\begin{figure}[h]
    \caption{Sensitivity Analysis: Missingness Type}
    \includegraphics[width = \textwidth, trim = {0 0 0 -0.1cm}, clip]{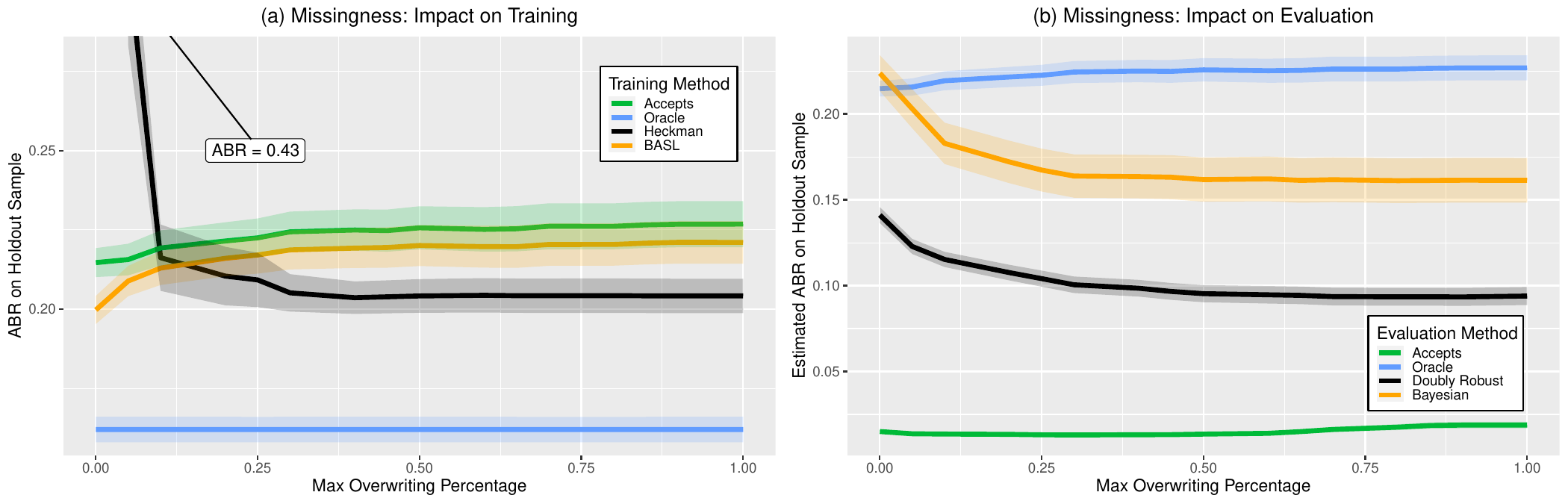}
    \label{fig_sensitivity_missingness}
    \footnotesize \textit{Note: }The figure depicts the relationship between overwriting and the impact of sampling bias on scorecard training and evaluation, benefits from our propositions (BASL and Bayesian evaluation) and benchmarks (Heckman model and doubly robust evaluation). The vertical axis of panel (a) is truncated at ABR = 0.40 for readability.
\end{figure}

We obtain the previous results on data generated according to a MAR process, where all applicants' features are revealed to the scorecard. Next, we introduce MNAR by generating data with three explanatory features and using only the first two in the scorecard, creating an omitted variable bias. On each iteration of the acceptance loop, we overwrite a certain percentage of the scorecard-based acceptance decisions by replacing applicants with the lowest values of the hidden feature. This allows us to gradually increase the strength of the MNAR process in the synthetic data. Figure \ref{fig_sensitivity_missingness} illustrates the loss due to sampling bias and the performance of our propositions depending on the overwriting percentage. We also implement two established bias correction benchmarks. BASL is compared to the Heckman model, which is developed for MNAR settings, whereas the Bayesian framework is compared to DR evaluation. 

We observe adverse effects of sampling bias on the training and evaluation of scorecards at any level of overwriting. This confirms that bias correction is required regardless of whether the data exhibits MAR or MNAR. In both extremes with no or full manual overwriting, loss in the ABR exceeds $.05$ during training. The impact on evaluation is stronger with a mean difference of $.21$ between the ABR estimated on accepts and the actual ABR on holdout applications. 

Our propositions consistently outperform accepts-based training and evaluation. Panel (a) of Figure \ref{fig_sensitivity_missingness} reveals the poor performance of the Heckman model under MAR. We explain this result by a high correlation between the two variables considered in the model: outcome (i.e., whether the applicant is \textit{bad}) and selection (i.e., whether the applicant was accepted). Previous studies also noted a poor Heckman performance in the presence of strong target-missingness correlation \citep{chen_economic_2001}. In our context, the correlation is high when scorecards are accurate. Traditional banks that focus on prime customers often obtain low default rates, which indicates the high accuracy of their scorecards. Increasing the overwriting percentage reduces the target-missingness correlation and strengthens the magnitude of the MNAR process, facilitating a superior performance of the Heckman model. In our simulations, Heckman outperforms BASL when more than $20\%$ of the applications undergo overwriting. In practice, such a large degree of manual intervention seems implausible for data-driven FIs, which increasingly rely on the automation of approval decisions but might be realistic for some banks, which would then be better served by using a Heckman-style model. However, since the type of missingness is difficult to identify in practice, the consistent superiority of BASL over the accepts-based benchmark leads us to conclude that institutions adopting BASL will not run the risk of impeding scorecard performance; which they might when using the Heckman model.  

Considering model evaluation, we observe the accuracy of the ABR estimates from the Bayesian framework to decrease in the MNAR setting. However, they are much closer to the true scorecard performance than estimates coming from accepts-based evaluation and DR across all setups. DR improves over ignoring rejects but still provides ABR estimates that are, on average, more than $.10$ off the actual model performance. This large gap can be explained by the adjustments required to apply DR in credit scoring, which we discuss in Appendix F.2.

%
%

\subsection{Real Data}
\label{sec_results_real}



\subsubsection{Experiment I: Evaluation with the Bayesian Framework}
Table \ref{tab_results_real_exp1} compares the accuracy of different model evaluation strategies. We compute the RMSE between the scorecard performance estimates produced by each evaluation strategy and the actual scorecard performance on the holdout sample representing the true borrower population.

\begin{table}[h]
    \caption{Scorecard Evaluation: Comparing Performance of Bias Correction Methods}
    \label{tab_results_real_exp1}
    {

\begin{tabular}{@{\extracolsep{9pt}} lccccc}
    \hline
    Evaluation method & AUC & BS & PAUC & ABR & Rank \\
    \hline 
    
    Ignore rejects               & .1234 (.0309) & .0306 (.0034) & .0983 (.0246) & .0356 (.0603)  & 2.46 \\
    Reweighting                  & .1277 (.0601) & .0348 (.0054) & .0826 (.3058) & .0315 (.0903)  & 2.49 \\
    Doubly robust                & --            & .0506 (.0050) & --            & .1167 (.0216)  & -- \\
    
    \textbf{Bayesian evaluation} & \textbf{.0111} (.1158) & \textbf{.0073} (.0213) & \textbf{.0351} (.0628) & \textbf{.0130} (.0331)  & \textbf{1.06} \\
    \hline
\end{tabular}}
    \begin{tablenotes}
      \small
      \item Values indicate RMSE between the actual scorecard performance on the holdout sample and estimates of that performance using an evaluation method. Variance of the performance estimates $\times 10^{-5}$ in parentheses.
    \end{tablenotes}
\end{table}

In line with the synthetic data results, we observe a relatively high RMSE when ignoring rejects, which evidences the loss due to sampling bias. 
Overoptimistic estimates of scorecard performance from the accepts-based evaluation lead to sub-optimal decisions. Expecting a certain default rate upon scorecard deployment, an FI would face losses and potential liquidity problems when encountering a substantially higher default rate among approved loans.

Weighted validation improves the accuracy of scorecard performance estimates for two evaluation metrics. Overall, reweighting performs marginally worse than accepts-based evaluation, achieving an average rank of $2.49$ compared to $2.46$. At the same time, reweighting outperforms accepts-based evaluation in the metrics that account for asymmetric error costs, PAUC, and ABR, which are most important for decision-makers. DR demonstrates a poor RMSE for the two supported metrics, BS and ABR. This can be attributed to the high difficulty of reward prediction in a high-dimensional environment and the limitations of DR when applied to credit scoring. Poor performance in the BS and ABR while lacking support for rank-based indicators such as the AUC and PAUC make DR an inappropriate evaluation method for the considered data set.

The Bayesian evaluation framework provides the most accurate estimates of the scorecard performance across all evaluation metrics and achieves an average rank of $1.06$. This implies that Bayesian evaluation produces the most reliable predictions of scorecard performance on new loan applications, helping decision-makers to anticipate the accuracy of a scorecard and judge its (business) value ex-ante. Appendix D.3 augments Table \ref{tab_results_real_exp1} with results from statistical testing. Pairwise Nemenyi post-hoc tests indicate that performance estimates obtained with the Bayesian framework are significantly better than those obtained with benchmark strategies at a 5\% level.


\subsubsection{Experiment II: Reject Inference with BASL} 
Table \ref{tab_results_real_exp2} compares the performance of bias correction methods. Notably, only four approaches achieve a lower rank than ignoring rejects. Labeling rejects as \textit{bad} performs worst. Given a historical acceptance rate of $20-40\%$ at Monedo, the underlying assumption of all rejects being \textit{bad} is too strong for this data. The bias-removing autoencoder also performs poorly. As discussed in Appendix F.3, due to a large number of features and a broad set of meta-parameters, the reconstruction error of the autoencoder remains high even after much tuning. This evidences the difficulty of using an autoencoder in high-dimensional settings.

\begin{table}[h]
    \caption{Scorecard Training: Comparing Performance of Bias Correction Methods}
    \label{tab_results_real_exp2}

\begin{tabular}{@{\extracolsep{3pt}} lcccccc}
     \hline
      Training method & AUC & BS & PAUC & ABR & Rank \\
     \hline      
     Ignore rejects            & .7984 (.0010) & .1819 (.0004) & .6919 (.0010) & .2388 (.0019) & 4.12 \\ 
     Label all as bad          & .6676 (.0014) & .2347 (.0006) & .6384 (.0010) & .3141 (.0022) & 9.56 \\ 
     Bias-removing autoencoder & .7304 (.0011) & .2161 (.0004) & .6376 (.0019) & .3061 (.0036) & 8.77 \\ 
     Bivariate probit          & .7444 (.0011) & .2124 (.0006) & .6397 (.0010) & .3018 (.0013) & 8.53 \\ 
     Bureau score based labels & .7978 (.0009) & .1860 (.0003) & .6783 (.0010) & .2514 (.0021) & 5.69 \\ 
     Hard cutoff augmentation  & .8033 (.0010) & .1830 (.0006) & .6790 (.0011) & .2458 (.0021) & 4.84 \\ 
     Two-step Heckman model    & .8034 (.0011) & .1828 (.0004) & .6934 (.0014) & .2381 (.0019) & 4.03 \\ 
     Reweighting               & .8040 (.0005) & .1840 (.0002) & .6961 (.0009) & .2346 (.0015) & 3.91 \\ 
     Parceling                 & .8038 (.0011) & .1804 (.0004) & .6885 (.0011) & .2396 (.0019) & 3.83 \\ 
     \textbf{Bias-aware self-learning} & \textbf{.8166} (.0007) & \textbf{.1761} (.0003) & \textbf{.7075} (.0011) & \textbf{.2211} (.0012) & \textbf{1.72} \\ 
     \hline
 \end{tabular}
    \begin{tablenotes}
      \small
      \item Values in parentheses represent standard errors across 100 trials of $4-$fold cross-validation $\times 25 $  bootstrap samples.
    \end{tablenotes}
\end{table}

The bivariate probit model improves on the previous benchmarks but performs worse than ignoring rejects. Its poor accuracy can be attributed to two reasons. First, a parametric probit model faces difficulties in handling high-dimensional and noisy data. To address this, we explore multiple variants of Heckman-style models described in Appendix F.4 that use different feature subsets, estimation procedures, and base classifiers. The best-performing two-step model presented in Table \ref{tab_results_real_exp2} uses XGB to model the outcome process and outperforms bivariate probit. Second, in line with the synthetic data results, the Heckman model performs poorly when the outcome and selection equations are highly correlated, which is typical under MAR or when the previous scorecard is accurate. Although it is infeasible to estimate the strength of the MNAR process on the real data, the poor performance of Heckman suggests that the missingness type is more geared towards MAR. A fully automated loan approval pipeline at Monedo further supports this view.

Considering model-based augmentation techniques, HCA performs on par with a two-step Heckman model, improving on ignoring rejects only in the AUC, whereas parceling performs better in three evaluation measures. The better performance of parceling can be explained by introducing randomness in the labeling stage, which helps reduce error propagation.

Reweighting outperforms other benchmarks in the AUC, PAUC and ABR. However, it performs worse than parceling in terms of BS, indicating poor calibration of the resulting scorecard. This translates to a marginally higher overall rank of reweighting compared to parceling. Appendix F.1 discusses the performance of different reweighting variants in detail, whereas Table \ref{tab_results_real_exp2} only includes the best-performing specification.

Independent of the performance indicator, BASL performs best and achieves the lowest average rank of $1.72$. Compared to reweighting, the closest competitor in the cost-sensitive metrics, the PAUC and ABR of the scorecard after bias correction with BASL increase by $.0114$ and $.0135$, respectively. Gains from BASL are statistically significant: Nemenyi post-hoc tests indicate that BASL significantly outperforms all benchmarks at a 5\% level in the AUC, PAUC, and ABR. Table D.6 in Appendix D provides ablations, which examine incremental performance gains from different BASL stages. While all stages add value, the largest gains stem from the filtering step. 

Experiment I and II confirm that our propositions outperform previous bias correction methods on the employed high-dimensional FinTech data set. To test external validity, we repeat experiments using gradually smaller subsets of features, simulating the transition from FinTechs to traditional banks, which tend to use parsimonious scorecards. The results suggest that BASL and Bayesian evaluation consistently reduce sampling bias across all settings, with larger gains observed on higher-dimensional data. The corresponding results are available in Appendix D.5.


\subsubsection{Business Impact Analysis.}
\label{subsubsec_impact}

To evaluate the business impact of our propositions, we strive to estimate monetary gains from BASL and Bayesian evaluation. This requires knowledge of key loan parameters. We consider the microloan market and draw loan amounts and interest rates from Gaussian distributions. In line with statistics from the US consumer loan market \citep{pew2016}, we assume a mean principal $A = \$375$ and a mean interest $i = 17.33\%$. 

In the event of default occurring with a probability $\mbox{PD}$, a financial institution recovers $A \times (1 + i) \times (1 - \mbox{LGD})$. If there is no default, the expected revenue is $A \times (1 + i)$. We approximate the loan-level PD by computing the ABR of the scorecard within a specified acceptance range. Using the modeling pipeline described in Section \ref{sec_setup}, we obtain $100$ ABR estimates for each scorecard. Given these 100 estimates, Equation \ref{eq_business_impact} yields an estimate of the average profit per loan:
\begin{equation}
\label{eq_business_impact}
\pi = \frac{1}{100} \sum_{j=1}^{100} \big[ \mbox{PD}_j \times A \times (1 + i) \times (1 - \mbox{LGD}) + (1 - \mbox{PD}_j) \times A \times (1 + i) - A \big]
\end{equation}

We aggregate the average profit per loan over 10,000 trials, drawing $A$ and $i$ from Gaussian distributions and varying LGD from $0$ to $1$. By subtracting the profit of each bias correction method from the profit of an accept-based scorecard, we obtain the incremental profit compared to ignoring rejects (and sampling bias). Finally, we compute the expected margin (i.e., the expected return per dollar issued) by dividing the incremental profit by the average loan amount. We perform these steps for BASL and reweighting, which is the strongest benchmark in terms of the ABR. Following the historical approval rates at Monedo, we vary the acceptance rates in $20-40\%$.  

Bayesian evaluation improves the accuracy of scorecard evaluation. To estimate gains from better evaluation, we consider a scenario in which a FI adjusts acceptance rates to maximize profit. We use Bayesian evaluation and benchmark methods to estimate the expected ABR of a scorecard across different acceptance rates in $10-50\%$, extending the range beyond the historical approval rates at Monedo. We use each evaluation strategy to select the acceptance rate that maximizes profit. Finally, we compare incremental profit under the acceptance rate selected based on the Bayesian evaluation, and the one based on reweighting.

\begin{figure}[t]
    \caption{Business Impact Analysis}
    \includegraphics[width = \textwidth, trim = {0 0 0 -0.1cm}, clip]{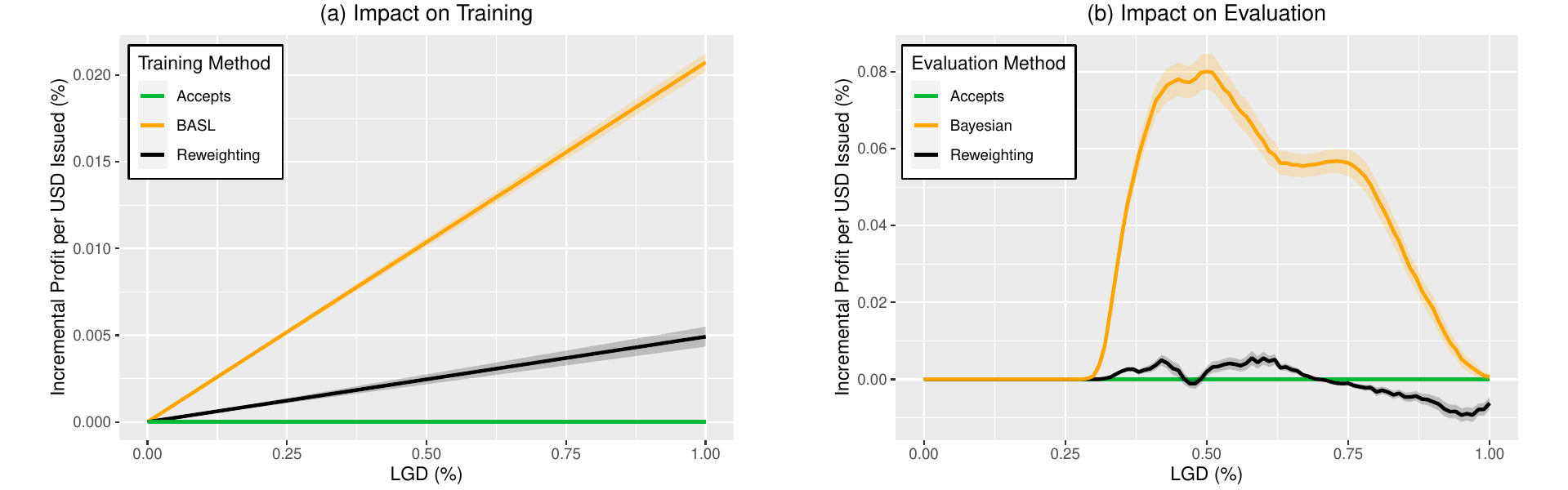}
    \label{fig_business_impact}
    \footnotesize \textit{Note: }The figure shows the expected incremental profit per dollar issued of our propositions and reweighting relative to a scorecard that ignores sampling bias. Panel (a) depicts gains from BASL on the scorecard training stage; panel (b) shows gains from Bayesian evaluation in a policy change scenario of selecting a new acceptance rate.
\end{figure}

Figure \ref{fig_business_impact} illustrates the expected incremental profit as a function of the LGD. In both settings, ignoring sampling bias impacts the profit of a financial institution. BASL increases the expected return per dollar issued by up to $2.07$ percentage points compared to ignoring rejects and up to $1.58$ compared to reweighting, whereby gains increase with the LGD as higher LGDs raise the cost of accepting a bad customer.

Bayesian evaluation increases the expected return per dollar issued by up to $5.67$ percentage points compared to both benchmarks. The gains are highest for LGD between $40\%$ and $80\%$ and diminish at very low or very high loss rates. At LGD below $30\%$, all evaluation strategies start recommending the maximal acceptance rate, as the cost of accepting a \textit{bad} customer becomes very small. Similarly, at LGD close to $100\%$ all evaluation strategies converge to recommending the minimal acceptance due to a substantially high cost of accepting a \textit{bad} customer. 

The business impact analysis demonstrates consistent monetary gains from the proposed bias correction methods. Assuming the average loan amount of $375\$$ and the mean LGD of $50\%$, the incremental profit per loan is $\$2.97$ for BASL and $\$29.27$ for Bayesian evaluation. This shows that correcting the impact of sampling bias on evaluation promises much larger financial rewards.

\section{Discussion}
\label{sec_discussion} 

Our propositions target the key steps in an ML workflow: training and evaluation. 
For scorecard training, simulation results suggest that sampling bias deteriorates scorecard performance, assessed in terms of the ABR, by, on average, six percentage points. Using BASL, we can recover about 25 percent of the loss due to bias. Results from the real-world data support the simulation and show BASL to consistently outperform several benchmarks by a 1.6 and up to 30 percent margin; depending on the accuracy indicator. According to the business impact analysis, these gains translate into an increase in the expected return per dollar between 1.5 to 2 percentage points compared to benchmarks and the ignoring rejects baseline.

The sensitivity analysis uncovers when BASL improves scorecard training and facilitates some policy prescriptions.
Observing gains from BASL to increase with decreasing acceptance rates and lower rates of good clients among applicants, we recommend using BASL for prime and/or non-collateralized financial products that FIs offer selectively. Building on supervised ML, BASL requires some defaults in the training data to infer a relationship between applicant characteristics and repayment behavior. More specifically, we observe the largest gains from BASL when the imbalance ratio is between 20 and 50, implying a bad ratio of two to five percent in the training data. Stronger class imbalance shrinks gains from BASL while more even class distributions imply more defaults among the accepted clients, which diminishes sampling bias and the need for RI. 

Using the real data, we study how the impact of sampling bias on scorecard accuracy varies with dimensionality, considering parsimonious scorecards with a handful of characteristics to scorecards with thousands of features. We find BASL to be most effective in high-dimensional environments. In lower dimensional settings, improvements are modest but consistent. 
These findings speak to the trend toward leveraging novel data sources for credit scoring 
\citep[e.g.,][]{Djeundje_2021_alternative_data}. 
Higher dimensionality may amplify sampling bias. Likewise, we conjecture that especially data-driven FIs, which use a sophisticated scoring methodology encompassing many characteristics, will benefit from BASL. We also expect FIs that have already deployed advanced scoring systems to be particularly alert about predictive performance, which would raise the acceptance of scoring innovations like BASL among managers and users.  

The missing data mechanism – although not identifiable in real life  - has crucial implications for RI \citep{banasik_reject_2007}. We elaborate on drivers for MAR and MNAR in credit scoring and examine the performance of BASL when gradually increasing the strength of confounding in the simulation. The experiment finds BASL to outperform the well-known Heckman model, which is designed for MNAR, across a range of plausible operating conditions. We also find Heckmann-style models to yield extremely poor risk predictions when the missingness mechanism leans toward MAR. The Heckman benchmark may still outperform BASL by a small margin under extreme conditions, for example, when assuming that credit analysts manually overwrite more than 25 percent of scorecard recommendations. Overall, our results provide evidence in favor of using BASL to address sampling bias. It promises competitive and robust performance independent of the form of missingness and avoids the risk of catastrophic predictions. Alternatives like the Heckman model are preferable if analysts have reasons to suspect their organization's scoring practices cause available data to exhibit strong MNAR. 
     
Another insight from our analysis concerns the trade-off between sampling bias mitigation and predictive accuracy. The missing data literature and much prior work on RI advocate the removal of bias through estimating (missing) labels for all rejected clients. We observe a trade-off between debiasing and scorecard performance. Examining the corresponding Pareto frontiers, we find that labeling a small, carefully selected set of rejects improves scorecard performance. In contrast, noise in the estimated labels causes a fully debiased data set to yield weaker scorecards. The design of BASL accounts for this trade-off, which contributes to outperforming previous RI methods. 

Turning to model evaluation, we find that the adverse impact of sampling bias is greater than in training and that our Bayesian framework facilitates substantial improvements over accept-based evaluation. 
In the simulation, we find the latter to severely underestimate default risks, predicting the ABR of a scorecard to be, on average, 0.01 compared to a true value of, on average, 0.21. The Bayesian framework corrects deviations between true and predicted ABR by about 54 percent. It also shows an appealing tendency that larger errors, if they occur, come from underestimating performance, which implies more prudent risk estimates.
Results from the real-data experiment confirm accuracy gains from Bayesian evaluation. For our micro-loan setting, improvements in predicting ABR are close to 60 percent over the strongest competitor. We also observe substantial accuracy gains for other performance indicators.

Appraising the business value of these improvements is nontrivial because scorecard performance prediction is a sparsely researched field and reference practices are lacking. We consider a policy optimization setting in which competing approaches forecast lending profits across alternative acceptance policies. For our micro-lending context, we observe Bayesian evaluation to raise the expected return per dollar issued by up to 5.7 percent (depending on LGD). Assuming an average loan amount of 375\$ and an average LGD of 50\%, which reflect the business history of Monedo, Bayesian evaluation increases the incremental profit per loan by 29.27\$, which equates to an eight percent improvement in profit. 

Sensitivity analysis secures the efficacy of Bayesian evaluation. All experimental conditions (e.g., varying acceptance rates, different feature sets, missingness according to MAR vs. MNAR, etc.) provide strong evidence in favor of the Bayesian framework. The corresponding forecasts are consistently much more accurate than alternative approaches to scorecard evaluation, facilitating a clear recommendation. Whenever FIs have access to the data of rejected clients, they should leverage this data for scorecard evaluation using the Bayesian framework. 
    
Notably, this recommendation extends to the challenging MNAR case although the Bayesian framework uses biased scores to infer the labels of rejects. The sensitivity analysis supports this approach. Using imperfect scores from the current - biased - scorecard as a prior for the labels of rejects produces better results than ignoring rejects during evaluation; even if the scores embody a lot of noise. This finding indicates that Bayesian evaluation is not only effective but also easy to use. FIs can simply use the original score of an applicant or their current scorecard to set the prior, which is a key ingredient of the Bayesian framework. 

\section{Conclusion}
\label{sec_conclusion}
Scoring models aid decision-making in banking. Studying loan approval decisions, the paper illustrates the adverse effects of sampling bias on the training and evaluation of scoring models and proposes methodologies to address these effects. Bayesian evaluation leverages unbiased unlabeled data for estimating model performance. The BASL framework mitigates the loss due to bias by training scorecards on an augmented sample, which comprises selected rejects with an inferred label. Using real-world lending data including an unbiased sample of randomly accepted loans, we confirm the effectiveness of BASL and the Bayesian framework and demonstrate their superiority over benchmarks. A business impact analysis evidences sizeable improvements in profit, especially for Bayesian evaluation. Results from a simulation study augment those findings by identifying boundary conditions determining the effectiveness of our propositions. 

Our study has several implications for credit scoring research and practice. We show that widely used evaluation regimes are vulnerable to sampling bias and give misleading advice. The Bayesian framework anticipates the loss in accuracy due to bias and facilitates quantifying this effect before scorecard deployment. A more realistic and forward-looking performance evaluation facilitates improving operational acceptance decisions to raise profitability and can help to improve risk management practices in the credit industry. 

The paper also confirms that training on biased data impairs scorecards. FIs can use BASL with any supervised learning algorithm to reduce loss in model performance. Doubt as to whether RI is worthwhile prevails in the literature \citep[e.g.,][]{chen_economic_2001}. Reporting positive results from an unbiased evaluation sample, the paper speaks to this debate. RI is a hard problem. Financial rewards will not be excessive. However, the specific engineering of BASL facilitated consistent and material gains in this study. Improvements of the magnitude observed here in a core business process may be a deciding factor in highly competitive lending markets.

A potentially important caveat of BASL and Bayesian evaluation is that they require access to unlabeled data of rejected clients. Meeting this requirement in a credit context is nontrivial. FIs need to store data on applicants they have rejected, which poses thorny questions related to privacy and consumer protection. Balancing the interests of lenders to gather more data for improving processes such as loan approval and the interests of consumers for protection against privacy infringement is a major challenge in the digital economy. Quantifying the value of a specific type of data in a specific business context, the paper contributes a humble piece of empirical evidence to this societal debate, which may inform bank governance and regulatory authorities.

\section*{Acknowledgments}
The authors thank the data science team of Monedo for the many fruitful discussions, which have helped tremendously in improving this research. Special thanks go to Yiannis Gatsoulis for his invaluable comments and inputs, which were instrumental to the paper. Stefan Lessmann acknowledges financial support through the project ``AI4EFin AI for Energy Finance'', contract number CF162/15.11.2022, financed under Romania’s National Recovery and Resilience Plan, Apel nr.~PNRR-III-C9-2022-I8.

%
%





%
%



\section*{Supplementary material}

# Fighting the Sampling Bias: A Framework for Training and Evaluating Credit Scoring Models

## Online Appendix

The online appendix provides supplementary material. It comprises: (i) literature tables surveying the prior work on bias correction and empirical studies on reject inference in credit scoring; (ii) methodological details on Bias-aware self-learning (BASL); (iii) extended empirical results on synthetic and real-world data; (iv) meta-parameters of the data generation process and bias correction methods; (v) implementation details and extended results for additional benchmarks. The appendix sections are independent and focus on specific topics. We encourage readers to read the table of contents and consult selected sections of interest.

## Contents

## Appendix A:   Prior Work on Bias Correction

This appendix includes the literature tables that provide a comprehensive overview of bias correction methods suggested in different research streams and summarize previous empirical studies on reject inference in credit scoring. A detailed description of the prominent bias correction methods is provided in Section 3.

### A.1.   Bias Correction Methods

Table A.1 overviews sampling bias correction methods suggested in the prior work. The bias correction methods suggested in different research streams are grouped into three families depending on the application stage: data preprocessing, model training and model evaluation. Data preprocessing methods encompass representation change techniques that transform input data before modeling to reduce the bias between the source and target distributions. Methods that correct sampling bias in the training stage split into two subgroups: model-based and reweighting techniques. Model-based techniques are embedded in a learning algorithm and account for the bias during model training by adjusting the optimization problem. Reweighting methods rebalance the loss function of a learning algorithm towards more representative data examples and can be applied during model training and model evaluation. Apart from reweighting, methods that correct bias in the evaluation stage include multiple evaluation metrics that approximate the generalization error under sampling bias and metric-agnostic evaluation frameworks.

In addition to the method type and the application stage, Table A.1 indicates two further characteristics of the bias correction methods: (i) whether the method is model-agnostic and (ii) whether it requires input data transformation. The advantage of model-agnostic methods is their flexibility with respect to the base classifier. Methods that rely on input data transformation require training a scoring model on latent features, which may harm the comprehensibility and explainability of the model.

### A.2.   Applications in Credit Scoring

Table A.2 overviews empirical studies on bias correction in credit scoring. We group the studies by the type of the implemented bias correction technique(s) and distinguish the methods applied in the model training and model evaluation stage. The discussion of the findings in Table A.2 is available in Section 3.3.

The empirical studies on reject inference are summarized across multiple dimensions, including: (i) whether the employed data set includes a holdout sample representative of the population, (ii) whether the performance gains are measured in profit and (iii) the number of features in the data set. The first dimension illustrates the potential for an accurate estimation of gains from bias correction, which requires labeled applications rejected by a scorecard. The second dimension shows if improvements from bias correction are measured in terms of the monetary gains. The third dimension indicates the data set dimensionality. The importance of handling the high-dimensional feature spaces is rising in light of a growing market share of FinTechs that operate with large amounts of data from different sources (Sirignano and Giesecke 2019) and an increasing reliance of financial institutions on alternative data sources such as applicant's digital footprint, e-mail activity and others (e.g. Jagtiani and Lemieux 2019).

**Table A.1    Sampling Bias Correction Methods**

| Reference | Method | Type | DP | TR | EV | MA | NT |
|---|---|---|---|---|---|---|---|
| Blitzer et al. (2006) | Structural correspondence learning | RC | ✓ | | | ✓ | |
| Daumé III (2009) | Supervised feature augmentation | RC | ✓ | | | ✓ | |
| Saenko et al. (2010) | Supervised feature transformation | RC | ✓ | | | ✓ | |
| Gopalan et al. (2011) | Sampling geodesic flow | RC | ✓ | | | ✓ | |
| Gong et al. (2012) | Geodesic kernel flow | RC | ✓ | | | ✓ | |
| Caseiro et al. (2015) | Unsupervised feature transformation | RC | ✓ | | | ✓ | |
| Satpal and Sarawagi (2007) | Penalized feature selection | RC | ✓ | | | ✓ | |
| Pan et al. (2011) | Transfer component analysis | RC | ✓ | | | ✓ | |
| Long et al. (2014b) | Transfer joint matching | RC | ✓ | | | ✓ | |
| Sun et al. (2016) | Correlation alignment | RC | ✓ | | | ✓ | |
| Wang and Rudin (2017) | Extreme dimension reduction | RC | ✓ | | | ✓ | |
| Atan et al. (2018) | Bias-removing autoencoder | RC | ✓ | | | ✓ | |
| Heckman (1979) | Heckman's model | MB | | ✓ | | | ✓ |
| Meng and Schmidt (1985) | Heckman-style bivariate probit | MB | | ✓ | | | ✓ |
| Lin et al. (2002) | Modified SVM | MB | | ✓ | | | ✓ |
| Daume III and Marcu (2006) | Maximum entropy genre adaptation | MB | | ✓ | | | ✓ |
| Yang et al. (2007) | Adapt-SVM | MB | | ✓ | | | ✓ |
| Marlin and Zemel (2009) | Multinomial mixture model | MB | | ✓ | | | ✓ |
| Bickel et al. (2009) | Kernel logistic regression | MB | | ✓ | | | ✓ |
| Chen et al. (2011) | Co-training for domain adaptation | MB, RC | ✓ | ✓ | | ✓ | |
| Duan et al. (2012) | Domain adaptation machine | MB | | ✓ | | | ✓ |
| Long et al. (2014a) | Regularized least squares | MB | | ✓ | | | ✓ |
| Liu and Ziebart (2014) | Robust bias-aware classifier | MB | | ✓ | | | ✓ |
| Joachims et al. (2017) | Modified ranking SVM | MB | | ✓ | | | ✓ |
| Chen et al. (2016) | Robust bias-aware regression | MB | | ✓ | | | ✓ |
| Liu et al. (2017) | Modified bias-aware classifier | MB | | ✓ | | | ✓ |
| Kügelgen et al. (2019) | Semi-generative model | MB | | ✓ | | | ✓ |
| Rosenbaum and Rubin (1983) | Model-based probabilities | RW | ✓ | ✓ | ✓ | ✓ | ✓ |
| Shimodaira (2000) | Distribution density ratios | RW | ✓ | ✓ | ✓ | ✓ | ✓ |
| Zadrozny (2004) | Selection probabilities are known | RW | ✓ | ✓ | ✓ | ✓ | ✓ |
| Huang et al. (2006) | Kernel mean matching | RW | ✓ | ✓ | ✓ | ✓ | ✓ |
| Cortes et al. (2008) | Cluster-based frequencies | RW | ✓ | ✓ | ✓ | ✓ | ✓ |
| Sugiyama et al. (2007b) | Kullback-Leibler weights | RW | ✓ | ✓ | ✓ | ✓ | ✓ |
| Kanamori et al. (2009) | Least-squares importance fitting | RW | ✓ | ✓ | ✓ | ✓ | ✓ |
| Loog (2012) | Nearest-neighbor based weights | RW | ✓ | ✓ | ✓ | ✓ | ✓ |
| Gong et al. (2013) | Focusing on cases similar to test data | RW | | ✓ | ✓ | ✓ | ✓ |
| Shimodaira (2000) | Modified AIC | EM | | | ✓ | | ✓ |
| Sugiyama and Ogawa (2001) | Subspace information criterion | EM | | | ✓ | | ✓ |
| Sugiyama and Müller (2006) | Generalization error | EM | | | ✓ | | ✓ |
| Sugiyama et al. (2007a) | Importance-weighted validation | EF | | | ✓ | ✓ | ✓ |
| Bruzzone and Marconcini (2010) | Circular evaluation strategy | EF | | | ✓ | ✓ | ✓ |
| **This paper** | **BASL and Bayesian evaluation** | **MB, EF** | | ✓ | ✓ | ✓ | ✓ |

Method types: RC = representation change, MB = model-based, RW = reweighting, EM = evaluation metric, EF = evaluation framework. Applications stages: DP = preprocessing, TR = training, EV = evaluation. Other abbreviations: MA = model-agnostic method, NT = does not involve input data transformation.

Table A.2    Empirical Studies on Reject Inference in Credit Scoring

| Reference | Implemented technique(s) | Training method(s) | Evaluation method(s) | Representative holdout sample | Profit gains | No. features |
|---|---|---|---|---|---|---|
| Joanes (1993) | Reclassification | DA | – | | | 3 |
| Fogarty (2006) | Multiple imputation | DA | – | | | 10 |
| Xia (2019) | Outlier detection with isolation forest | DA | – | | | 9 |
| Lin et al. (2020) | Ensembling classifiers and clusters | MB | – | | | 5, 23 |
| Kang et al. (2021) | Label spreading with oversampling | DA | – | | | 22 |
| Boyes et al. (1989) | Heckman model variant (HM) | MB | – | | | 42 |
| Feelders (2000) | Mixture modeling | MB | – | | | 2 |
| Chen and Astebro (2001) | HM | MB | – | ✓ | | 24 |
| Banasik et al. (2003) | HM | MB | – | ✓ | | 30 |
| Wu and Hand (2007) | HM | MB | – | | | 2 |
| Kim and Sohn (2007) | HM | MB | – | ✓ | | 16 |
| Chen and Astebro (2012) | Bayesian model | MB | – | | ✓ | 40 |
| Li et al. (2017) | Semi-supervised SVM (S3VM) | MB | – | | | 7 |
| Marshall et al. (2010) | HM | MB | – | | | 18 |
| Tian et al. (2018) | Kernel-free fuzzy SVM | MB | – | | | 7, 14 |
| Xia et al. (2018) | CPLE-LightGBM | MB | – | | | 5, 17 |
| Anderson (2019) | Bayesian network | MB | – | | | 7, 20 |
| Kim and Cho (2019) | S3VM with label propagation | MB | – | | | 17 |
| Shen et al. (2020) | Unsupervised transfer learning | MB | – | | | 20 |
| Banasik and Crook (2005) | Banded weights | RW | – | ✓ | | 30 |
| Verstraeten and Van den Poel (2005) | Resampling | RW | – | ✓ | | 45 |
| ? | Missing data based weights | RW | – | | ✓ | 40 |
| Crook and Banasik (2004) | Banded weights, extrapolation | RW, DA | – | ✓ | | 30 |
| Banasik and Crook (2007) | HM with banded weights | MB, RW | – | ✓ | | 30 |
| Maldonado and Paredes (2010) | Self-learning, S3VM | MB, DA | – | | | 2, 20, 21 |
| Anderson and Hardin (2013) | HCA, Mixture modeling | DA, MB | – | | | 12 |
| Nguyen (2016) | Parceling, HM, Banded weights | DA, MB, RW | – | | | 9 |
| Mancisidor et al. (2020) | Bayesian model, self-learning, S3VM | DA, MB | – | | | 7, 58 |
| **This paper** | **BASL, Bayesian evaluation** | **DA** | **EF** | ✓ | ✓ | **2,410** |

Abbreviations: DA = data augmentation, MB = model-based, RW = reweighting, EF = evaluation framework. "Representative holdout sample" indicates whether the study has access to a sample from the borrower's population for evaluation. "Profit gains" indicates whether gains from bias correction are measured in terms of profit. "Heckman model variant" includes a linear Heckman model and a bivariate probit/logistic model with non-random sample selection.

## Appendix B: Bias-Aware Self-Learning Framework

This appendix consists of two parts: (i) it provides the pseudo-code describing the BASL reject inference framework; (ii) it elaborates on the benefits of using a weak learner such as LR to label rejected applications.

### B.1. Framework Pseudo-Code

BASL includes four stages: (i) filtering rejects, (ii) labeling rejects, (iii) training the scorecard, (iv) early stopping. Algorithm B.1 provides the pseudo-code describing the filtering stage. Algorithm B.2 describes the labeling stage. The complete BASL framework is summarized in Algorithm B.3 and explained in Section 5.

### B.2. Labeling Rejects with a Weak Learner

This section illustrates the importance of using a weak learner on the labeling stage of the BASL framework. Consider two scoring models that employ different base classifiers: (i) $f_{XGB}(X)$ uses a strong non-parametric learner (XGB); (ii) $f_{LR}(X)$ uses a weak learner (L1-regularized LR). Using a real-world data set described in Section 6, we train both models on $D^a$ and use them to score applications in the holdout sample $H$.

Figure B.1 compares the distribution densities of predictions of $f_{XGB}$ and $f_{LR}$. LR produces probabilistic predictions that follow a distribution with a high kurtosis and a density peak around .40. In contrast, XGB scores follow a distribution with heavier tails. The range of the scores of $f_{XGB}$ is wider than that of $f_{LR}$, and the predictions are close to extreme values of 0 or 1 more frequently. This indicates a poor calibration of boosting predictions, which is also confirmed by the previous studies (Niculescu-Mizil and Caruana 2005).

---

**input** : accepts $\mathbf{X}^a$, rejects $\mathbf{X}^r$, meta-parameters $\beta = (\beta_l, \beta_u)$

**output:** filtered rejects $\mathbf{X}^r$

1   $g(X) =$ novelty detection algorithm trained over $\mathbf{X}^a$ ;

2   $\mathbf{s}^r = g(\mathbf{X}^r)$ ;                          `// predict similarity scores`

3   $\mathbf{X}^r = \{X_i \in \mathbf{X}^r | s_i^r \in [\beta_l, \beta_u] \}$, $\beta_l$ and $\beta_u$ are percentiles of $\mathbf{s}^r$ ;      `// filter rejects`

    **return :** $\mathbf{X}^r$

**Algorithm B.1:** Bias-Aware Self-Learning: Filtering Stage

---

**input** : labeled accepts $D^a = \{(\mathbf{X}^a, \mathbf{y}^a)\}$, unlabeled rejects $D^r = \{\mathbf{X}^r\}$, meta-parameters $\rho, \gamma, \theta$

**output:** selected labeled rejects $D^* = \{(\mathbf{X}^*, \mathbf{y}^*)\}$

1   $\mathbf{X}^* = \text{sample}(\mathbf{X}^r, \rho)$, $\rho$ is the sampling rate ;             `// random sample of rejects`

2   $f(X) =$ weak learner trained over $\mathbf{D}^a$ ;

3   $\mathbf{s}^* = f(\mathbf{X}^*)$ ;                        `// score rejects with a weak learner`

4   derive $c_g$: $\mathbf{P}(\mathbf{s}^* < c_g) = \gamma$, $\gamma$ is the percentile threshold ;

5   derive $c_b$: $\mathbf{P}(\mathbf{s}^* > c_b) = \gamma\theta$, $\theta$ is the imbalance parameter ;

6   $\mathbf{X}^* = \{X_i^* \in \mathbf{X}^* | s_i^* < c_g \text{ or } s_i^* > c_b\}$ ;           `// select relevant examples`

7   $\mathbf{y}^*$: $y_i^* = 0$ if $s_i^* < c_g$ and $y_i^* = 1$ if $s_i^* > c_b$ ;         `// assign labels`

    **return :** $\{(\mathbf{X}^*, \mathbf{y}^*)\}$

**Algorithm B.2:** Bias-Aware Self-Learning: Labeling Stage

---

**input** : labeled accepts $D^a = \{(\mathbf{X}^a, \mathbf{y}^a)\}$, unlabeled rejects $D^r = \{\mathbf{X}^r\}$, holdout set $H = \{\mathbf{X}^h\}$;

meta-parameters: filtering ($\beta$), labeling ($\tau, \gamma, \theta$), stopping criteria ($j_{max}$, $\mathbf{P}(y^r|\mathbf{X}^r), \epsilon$)

**output:** corrected scoring model $f_c(X)$

**1** $\mathbf{X}^r = \text{filtering}(\mathbf{X}^a, \mathbf{X}^r, \beta)$ ;                                       // filtering stage

**2** $j = 0$; $V = \{\}$; $F = \{\}$ ;                                                          // initialization

**3** **while** $(j \leq j_{max})$ *and* $(\mathbf{X}^r \neq \emptyset)$ *and* $(V_j \geq V_{j-1})$ **do**

**4**     $j = j + 1$

**5**     $D^* = \{(\mathbf{X}^*, \mathbf{y}^*)\} = \text{labeling}(D^a, D^r, \rho, \gamma, \theta)$ ;                      // labeling stage

**6**     $D^a = D^a \cup D^*$ ;                               // append labeled rejects to the labeled sample

**7**     $\mathbf{X}^r = \mathbf{X}^r - \mathbf{X}^*$ ;                        // remove labeled rejects from the unlabeled sample

**8**     $F_j = f(X) = \text{strong learner trained over augmented } D^a$ ;                     // training stage

**9**     $V_j = \text{BM}(F_j, H, \mathbf{P}(y^r|X^r), \epsilon)$ ;                          // evaluate using a Bayesian metric

**10** **end**

**11** **return** $f_c(X) = F_{\arg\max(V)}$ ;                              // return best-performing strong learner

**Algorithm B.3:** Bias-Aware Self-Learning Framework

---

Within the BASL framework, we only label rejects for which the labeling model has high confidence in the predicted labels, focusing on the examples from the tails of the score distribution. This requires a labeling algorithm that is able to accurately assign labels for examples on the extremes of the score distribution and has a good calibration ability to facilitate setting the appropriate confidence thresholds based on the predicted probabilities of default. As indicated in Figure B.1, a weak learner such as LR is a good fit. A weak learner is also less prone to overfitting the biased training sample, which could increase the risk of error propagation during labeling. Furthermore, as explained in Section 5, a regression-type algorithm such as LR is able to extrapolate outside of the feature range observed in the training data, which is not the case for tree-based learners such as XGB. Therefore, we use LR on the labeling stage of BASL.

## Appendix C:   Extended Results on Synthetic Data

This appendix provides methodological details of the simulation framework introduced in Section 6 and additional empirical results that extend the results reported in Section 7.1. The simulation study illustrates sampling bias, its adverse effect on the behavior, training and evaluation of scoring models and gains from our two propositions: the Bayesian evaluation framework and BASL. The parameters of the data generation process and the acceptance loop used in the simulation are provided in Appendix E.

### C.1.   Simulation Framework

The simulation framework is summarized in Algorithm C.2. The framework consists of two stages: the initialization and the acceptance loop. In the initialization stage, we generate synthetic data including two classes of borrowers from a mixture of Gaussian distributions using Algorithm C.1. A similar approach to generate synthetic loan applications using Gaussian distributions has been used in prior work (e.g.,

Martens et al. 2007, Maldonado and Paredes 2010). Let our synthetic examples $\mathbf{X}^g = (X_1^g, ..., X_n^g)^\top$ and $\mathbf{X}^b = (X_1^b, ..., X_m^b)^\top$ representing *good* and *bad* loan applications be generated as follows:

$$\begin{cases} \mathbf{X}^g \sim \sum_{c=1}^C \delta_c \mathcal{N}_k(\boldsymbol{\mu}_c^g, \boldsymbol{\Sigma}_c^g) \\ \mathbf{X}^b \sim \sum_{c=1}^C \delta_c \mathcal{N}_k(\boldsymbol{\mu}_c^b, \boldsymbol{\Sigma}_c^b) \end{cases} \quad (1)$$

where $\delta_c$ is the weight of the $c$-th Gaussian function, $\sum_{c=1}^C \delta_c = 1$, and $\boldsymbol{\mu}_c$ and $\boldsymbol{\Sigma}_c$ are the mean vector and the covariance matrix of the $c$-th Gaussian. The elements of $\boldsymbol{\Sigma}_c^t$ are drawn from a uniform distribution $\mathcal{U}(0, 1)$. We also append two noisy features with the same mean and variance for both classes: $x_\varepsilon \sim \mathcal{N}(0, 1)$.

Suppose the random binary vector $\mathbf{y} = \mathbf{y}^g \cup \mathbf{y}^b$ is a label indicating if an applicant is a *good* ($y = 0$) or *bad* risk ($y = 1$). The difference between the applicant classes is controlled by the parameters of the underlying distributions. Assuming a *bad* rate of $b$, we generate $n_b = nb$ *bad* examples and $n_g = n(1 - b)$ *good* examples and construct a first batch of the loan applications $D^* = \{(\mathbf{X}^*, \mathbf{y}^*)\}$ with $(\mathbf{X}^*, \mathbf{y}^*) \sim \mathbf{P}_{XY}$. We also generate a holdout set of $h$ examples denoted as $H = \{(\mathbf{X}^h, \mathbf{y}^h)\}$ using the same parameters as for the initial population. $H$ acts as a representative set and does not exhibit sampling bias. We use $H$ for performance evaluation.

The second stage of the framework – the acceptance loop – simulates the dynamic acceptance process, where loan applications arrive in batches over certain periods of time (e.g., every working day). Assume that

**Figure B.1    Prediction Density Comparison**

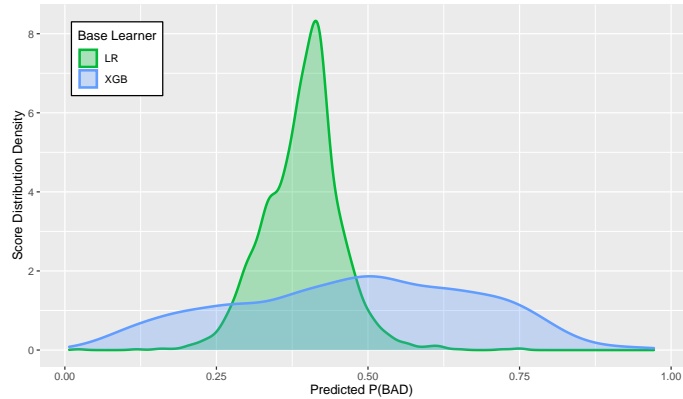

*Note.* The figure compares distribution densities of the scores predicted by two scoring models using different base classifiers: (i) L1-regularized logistic regression $f_{LR}$ (green); (ii) extreme gradient boosting $f_{XGB}$ (blue).

---

**input**  : distribution parameters $\boldsymbol{\mu}_c^g, \boldsymbol{\mu}_c^b, \boldsymbol{\Sigma}_c^g, \boldsymbol{\Sigma}_c^b, \delta_c, C$, sample size $n$, *bad* ratio $b$

**output**: labeled set of examples $D = \{(\mathbf{X}, \mathbf{y})\}$

**1** $n_b = bn; \, n_g = n - n_b$ ;                                 `// compute class-specific sample sizes`

**2** $\mathbf{X}^g \sim \sum_{c=1}^C \delta_c \mathcal{N}_k(\boldsymbol{\mu}_c^g, \boldsymbol{\Sigma}_c^g)$ ;                              `// generate $n_g$ good applications`

**3** $\mathbf{X}^b \sim \sum_{c=1}^C \delta_c \mathcal{N}_k(\boldsymbol{\mu}_c^b, \boldsymbol{\Sigma}_c^b)$ ;                              `// generate $n_b$ bad applications`

**4** $\mathbf{y}^g = \vec{0}; \, \mathbf{y}^b = \vec{1}$ ;                                      `// define applications' labels`

**5** $D = \{(\mathbf{X}^g, \mathbf{y}^g) \cup (\mathbf{X}^b, \mathbf{y}^b)\}$ ;                          `// construct a data set`

**return**: $D$

**Algorithm C.1:** Synthetic Data Generation

$D^* = \{\mathbf{X}^*\}$ is the first batch of $n$ applicants a financial institution encounters when entering a new market. Since no repayment data have been collected so far, a company might rely on a simple business rule to filter applications. An example would be to rank applications in $D^*$ by their credit bureau scores denoted as $x_v$. In our simulation, $x_v$ refers to the feature with the largest difference in mean values between *good* and *bad* applicants and represents a powerful attribute, such as a bureau score, which can be used to perform a rule-based application ranking. Assuming the target acceptance rate of $\alpha$, the financial institution grants a loan to the $\alpha n$ applicants with the highest bureau scores, forming a set of accepts $D^a = \{X_i \in \mathbf{X}^* | x_{i,v} \geq \tau\}$, and reject $(1-\alpha)n$ remaining applicants, forming a set of rejects $D^r = \{X_i \in \mathbf{X}^* | x_{i,v} \leq \tau\}$, where $\tau$ is the $(1-\alpha)$-th percentile of $x_v$ with respect to $D^*$. Eventually, the repayment status of applicants in $D^a$ is observed, providing the corresponding labels $\mathbf{y}^a$. The labeled set $D^a = \{(\mathbf{X}^a, \mathbf{y}^a)\}$ can now be used to train a scoring model $f_a(X)$ to support the acceptance decisions for the incoming loan applications.

On each iteration of the acceptance loop, $f_a(X)$ is trained over the available set of accepts in $D^a$. In addition to $f_a(X)$, we also train an oracle model $f_o(X)$ over the union of accepts and rejects $D^a \cup D^r$. The model $f_o$ represents a theoretical optimum as it uses representative data that is not available in practice and does not suffer from sampling bias. We use both $f_a$ and $f_o$ to score examples in $H$ and evaluate their performance. Next, we generate a batch of $n$ new applicants $D_j = \{(\mathbf{X}, \mathbf{y})\}$ using the same distribution parameters as for the initial population (i.e., assuming the absence of population drift) and predict the scores of the new applicants using $f_a$. Based on the model predictions, we accept $\alpha n$ examples with the lowest

---

> **input** : distribution parameters $\boldsymbol{\mu}_c^g, \boldsymbol{\mu}_c^b, \boldsymbol{\Sigma}_c^g, \boldsymbol{\Sigma}_c^b, \delta_c, C$, sample sizes $n, h$, *bad* ratio $b$, acceptance rate $\alpha$, number of iterations $j_{max}$, feature indicator $v$
>
> **output:** labeled accepts $D^a$, labeled rejects $D^r$, labeled holdout set $H$
>
> **1** $D^* = \{(\mathbf{X}^*, \mathbf{y}^*)\} = \text{generate}(\boldsymbol{\mu}_c^g, \boldsymbol{\mu}_c^b, \boldsymbol{\Sigma}_c^g, \boldsymbol{\Sigma}_c^b, \delta_c, C, b, n)$ ;  `// generate data using Algorithm C.1`
>
> **2** $H = \{(\mathbf{X}^s, \mathbf{y}^s)\} = \text{generate}(\boldsymbol{\mu}_c^g, \boldsymbol{\mu}_c^b, \boldsymbol{\Sigma}_c^g, \boldsymbol{\Sigma}_c^b, \delta_c, C, b, h)$ ;            `// generate holdout set`
>
> **3** $\tau = (1-\alpha)$-th percentile of $x_v$ with respect to $D^*$ ;            `// simple business rule`
>
> **4** $D^a = \{(X_i^*, y_i^*) | x_{i,v} \geq \tau\}$ ;            `// accept` $\alpha n$ `applications`
>
> **5** $D^r = \{(X_i^*, y_i^*) | x_{i,v} < \tau\}$ ;            `// reject` $(1-\alpha)n$ `applications`
>
> **6 for** $j \in \{1, 2, ..., j_{max}\}$ **do**
>
> **7** $\quad$ $f_a(X)$ = accepts-based model trained over $D^a$ ;
>
> **8** $\quad$ $f_o(X)$ = oracle model trained over $D^a \cup D^r$ ;
>
> **9** $\quad$ $D_j = \{(\mathbf{X}, \mathbf{y})\} = \text{generate}(\boldsymbol{\mu}_c^g, \boldsymbol{\mu}_c^b, \boldsymbol{\Sigma}_c^g, \boldsymbol{\Sigma}_c^b, \delta_c, C, b, n)$ ;            `// batch of new applications`
>
> **10** $\quad$ $\tau = \alpha$-th percentile of $f_a(D_j)$ ;            `// compute acceptance threshold`
>
> **11** $\quad$ $D_j^a = \{(X_i, y_i) | f_a(X_i) \leq \tau\}$ ;            `// accept` $\alpha n$ `applications`
>
> **12** $\quad$ $D_j^r = \{(X_i, y_i) | f_a(X_i) > \tau\}$ ;            `// reject` $(1-\alpha)n$ `applications`
>
> **13** $\quad$ $D^a = \bigcup_{i=1}^{j} D_i^a$; $D^r = \bigcup_{i=1}^{j} D_i^r$ ;            `// append accepts and rejects`
>
> **14 end**
>
> $\quad$**return :** $D^a, D^r, H$

**Algorithm C.2:** Simulation Framework

predicted scores and reject the remaining $(1 - \alpha)n$ applications. The newly rejected examples are appended to $D^r$, whereas the newly accepted examples and their labels are appended to $D^a$. The augmented set of accepts is used to retrain $f_a$ in the next iteration of the acceptance loop.

To illustrate performance gains from BASL, we apply it on each iteration of the acceptance loop. We train a scoring model $f_c(X)$ that has undergone bias correction by applying BASL to the available set of rejects $D^r$ and augmenting the training set $D^a$ with the selected labeled rejects. On each iteration, we train $f_c$ on the augmented data and score examples in $H$. This allows us to track gains from BASL compared to $f_a$.

Gains from the Bayesian evaluation framework are demonstrated by comparing the actual performance of $f_a$ on a representative holdout sample $H$ (labeled as oracle performance) and the predicted performance of $f_a$ estimated with different evaluation methods on each iteration of the acceptance loop. First, $f_a$ is evaluated on a validation subset drawn from the available set of accepts $D^a$. Second, we evaluate $f_a$ on a validation set that consists of the labeled accepts in $D^a$ and pseudo-labeled rejects in $D^r$ using the Bayesian evaluation framework. This allows us to quantify the gap between the actual and predicted performance of the scorecard and measure the recovery of this gap by using the Bayesian framework for performance evaluation.

## C.2. Experiment I

This section provides the extended results of Experiment I on synthetic data in the MAR setup. Table C.3 compares the performance of accepts-based evaluation and the Bayesian evaluation framework. The table quantifies the difference between the actual scorecard performance on a representative holdout set and the predicted scorecard performance estimated with one of the two evaluation strategies. We measure bias, variance and RMSE of the performance estimates using four evaluation metrics: the AUC, BS, PAUC and ABR. The results are aggregated across 100 simulation trials $\times$ 500 acceptance loop iterations.

According to Table C.3, performance estimates provided by the Bayesian evaluation framework have a lower bias than those obtained within the accepts-based evaluation. Despite accepts-based estimates demonstrating a lower variance in two evaluation metrics, the BS and ABR, RMSE values between the actual and predicted scorecard performance clearly indicate the advantage of using the Bayesian framework for scorecard evaluation. In all considered evaluation metrics, the Bayesian framework is able to provide a better estimate of the scorecard performance on unseen cases from the population of borrowers.

**Table C.3    Experiment I Results on Synthetic Data**

| Evaluation method | Metric | Bias | Variance | RMSE |
|---|---|---|---|---|
| Accepts-based evaluation | AUC | .1923 | .0461 | .2205 |
| Bayesian framework | | .0910 | .0001 | .1000 |
| Accepts-based evaluation | BS | .0748 | .0006 | .0828 |
| Bayesian framework | | .0038 | .0009 | .0566 |
| Accepts-based evaluation | PAUC | .2683 | .0401 | .2803 |
| Bayesian framework | | .1102 | .0002 | .1187 |
| Accepts-based evaluation | ABR | .1956 | .0004 | .2010 |
| Bayesian framework | | .0039 | .0040 | .0929 |

Abbreviations: AUC = area under the ROC curve, BS = Brier Score, PAUC = partial AUC on FNR $\in [0, .2]$, ABR = average *bad* rate among accepts at 20-40% acceptance, RMSE = root mean squared error.

## C.3.   Experiment II

Table C.4 presents the extended results of Experiment II on synthetic data in the MAR setup. The table provides the average loss due to sampling bias using five metrics. First, we use four scorecard performance metrics, the AUC, BS, PAUC and ABR, to measure the performance deterioration. The loss due to bias is measured as a difference between the performance of the oracle model trained on the union of accepts and rejects and that of the accepts-based model trained on accepts only. Second, we measure the loss in the MMD metric, which represents the magnitude of sampling bias in the labeled training data. The MMD is calculated between the training data of accepts and the representative holdout sample. The gains from reject inference with BASL are measured as a percentage from the corresponding loss due to sampling bias in each metric. The results are averaged across 100 simulation trials × 500 acceptance loop iterations.

The results suggest that the loss due to sampling bias is observed in all considered performance metrics. BASL consistently recovers between 22% and 36% of the loss. The largest performance gains are observed in the AUC and the BS, which represent the metrics that disregard error costs and are measured on the full set of credit applicants. The gains in the two cost-sensitive metrics measured on the subset of applications deemed as least risky, the PAUC and the ABR, are smaller but still exceed 22%. This suggests that gains from reject inference are observed through both type I and type II error reduction.

Interestingly, the results in the MMD metric indicate that augmenting the training data of accepts with rejects labeled by BASL improves the MMD by just 3.74%. This implies that the training data still exhibits a strong sampling bias. At the same time, using that data to train a corrected scoring model recovers more than 22% of the loss due to bias and scorecard performance, respectively. This result emphasizes that it is enough to label only a portion of rejected cases that help to improve the predictive performance and is further supported by the results of the accuracy-bias trade-off analysis provided in Appendix C.4. Increasing the number of labeled rejects allows to further improve the MMD, but does not lead to better scorecard performance due to the noise in the assigned labels. The trade-off between introducing too much noise in the labels and gains from having more representative training data is, therefore, a crucial part of BASL.

## C.4.   Bias-Accuracy Trade-Off

This section investigates the dependency between sampling bias in the data used for scorecard development and the accuracy of the trained scorecard in more detail. Using synthetic data, we compare multiple variants of BASL and reject inference techniques that perform data augmentation (i.e., label rejected applications

**Table C.4     Experiment II Results on Synthetic Data**

| Metric | Loss due to bias | Gain from BASL |
|--------|------------------|----------------|
| AUC | .0591 | 35.72% |
| BS | .0432 | 29.29% |
| PAUC | .0535 | 22.42% |
| ABR | .0598 | 24.82% |
| MMD | .5737 | 3.74% |

Abbreviations: AUC = area under the ROC curve, BS = Brier Score, PAUC = partial AUC on FNR $\in [0, .2]$, ABR = average *bad* rate among accepts at 20-40% acceptance, MMD = maximum mean discrepancy.

and append them to the training data). The results demonstrate the importance of limiting the number of labeled rejects to obtain the best performance and indicate a trade-off between performance maximization and bias mitigation.

The analysis is performed on the synthetic data, which we describe in detail in Appendix C.1. After running the acceptance loop, we apply different reject inference techniques to augment the biased training data of accepts and measure the accuracy and bias of each technique. First, we evaluate the performance of each reject inference method using the performance metrics considered in the paper: the area under the ROC curve (AUC), the partial AUC (PAUC), the Brier score (BS) and the average *bad* rate among accepts (ABR). Second, we evaluate the magnitude of sampling bias in the augmented training data after reject inference. Here, we use the maximum mean discrepancy metric (MMD, Borgwardt et al. 2006), which measures the similarity of the feature distribution between the augmented training data and the unbiased holdout sample.

We implement different variants of BASL, varying the meta-parameter values such that a different subset of rejects is selected in the labeling stage of the framework. This allows us to consider BASL variants that arrive at different points in the accuracy-bias space. Labeling more rejects facilitates reducing sampling bias, as the distribution mismatch between the training data and the holdout sample diminishes when adding more rejected applications. On the other hand, noise in the pseudo-labels assigned to the appended rejects harms the performance of the resulting scorecard. To study the trade-off between these conflicting dimensions, we construct Pareto frontiers that contain the non-dominated BASL solutions in the bias-accuracy space.

Our approach to measuring bias after reject inference is only suitable for data augmentation methods that label rejects and expand the training data. Some reject inference methods (e.g., the Heckman model) do not explicitly label rejects. Therefore, apart from BASL, our experiment includes the following benchmarks: ignoring rejects, labeling all rejects as *bad* risks, hard cutoff augmentation (HCA), and parceling. In addition to the Pareto frontiers with non-dominated BASL variants, we also depict some dominated BASL solutions with a high MMD to sketch the bias-accuracy relationship when labeling fewer rejects. For that purpose, we split the MMD interval between the non-dominated BASL variant with the highest MMD and ignoring rejects into equal bins and display the best-performing BASL variant within each of the bins. Figure C.2 demonstrates the results.

Figure C.2 shows that ignoring rejects leads to the strongest sampling bias in the training data since it only includes accepts, exhibiting MMD of around .60. The data augmentation benchmarks – labeling rejects as *bad*, HCA and parceling – completely eliminate sampling bias and reduce the MMD to around 0. This low MMD is achieved by labeling all rejects, which provides training data that represents the borrower population. However, a high reduction in the MMD does not necessarily improve the performance of the corrected scorecard. Except for the AUC, where all benchmarks improve on ignoring rejects, only some of the three data augmentation techniques outperform the scorecard that ignores rejects. This can be explained by the noise in the pseudo-labels of rejects that results from labeling all rejects, including those that are very different from the accepts.

The BASL framework includes multiple steps to restrict the labeling to selected rejects and attend to the distribution similarity between accepts and rejects and the model's confidence in the assigned label.

**Figure C.2    Bias-Accuracy Trade-Off of Reject Inference Techniques**

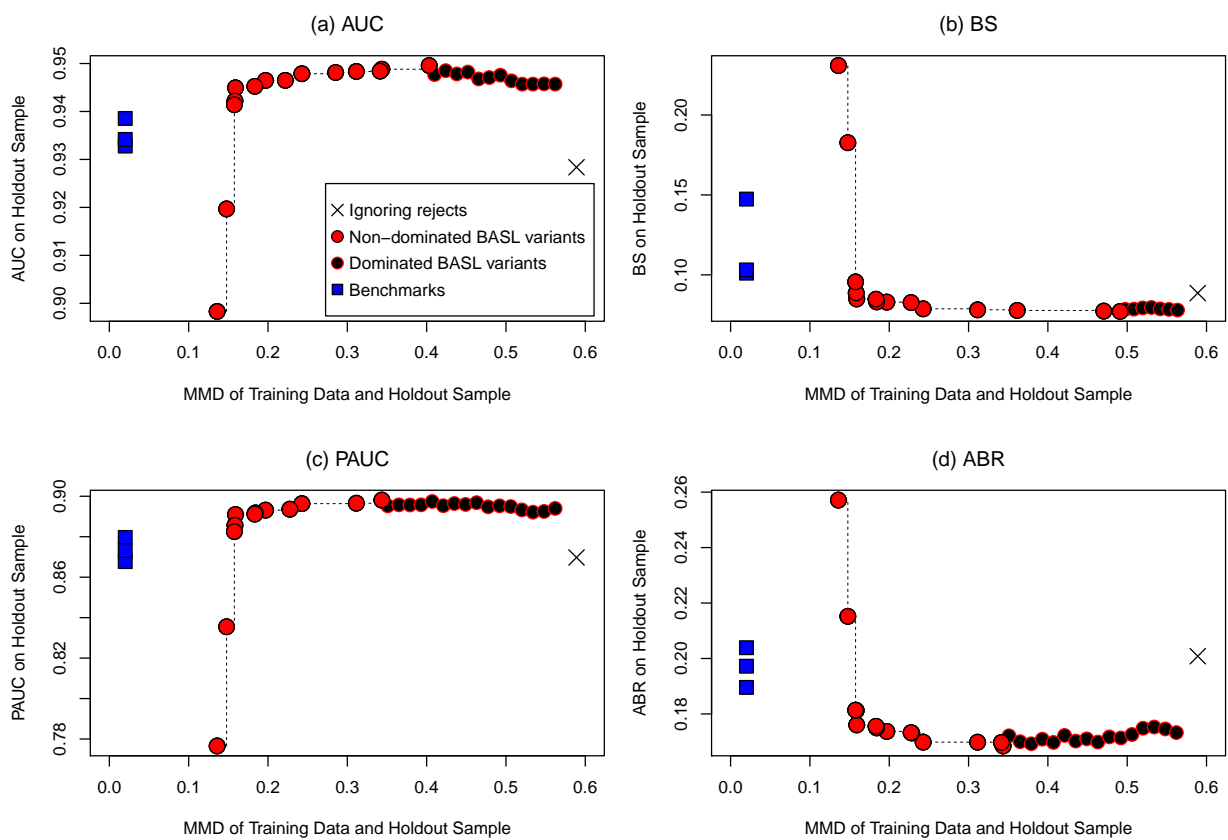

*Note.* The figure illustrates the trade-off between the scorecard accuracy and sampling bias when performing reject inference with data augmentation techniques. The vertical axes measure the scorecard performance in one of the four metrics: area under the ROC curve (AUC), Brier score (BS), partial AUC on FNR $\in [0, .2]$ (PAUC), average *bad* rate among accepts at 20-40% acceptance (ABR). The horizontal axes quantify sampling bias in the (augmented) training data by calculating the mean maximum discrepancy (MMD) between the training data and the representative holdout sample. The black cross refers to ignoring rejects. The blue squares depict data augmentation benchmarks that label all rejects: label all as *bad*, hard cutoff augmentation and parceling. The red points depict non-dominated BASL variants with different meta-parameter values. The non-dominated BASL variants label between 3% and 42% of rejects. The black points refer to the best dominated BASL variants that label between 1% and 3% of rejects.

Limiting the number of labeled rejects substantially decreases the gain in MMD. The BASL variants lying on the Pareto frontiers label between 3% and 42% of the rejects after multiple labeling iterations. This allows decreasing the MMD to some value in the range between .40 and .15, indicating that the training data still exhibits sampling bias. We obtain the best performance from scorecards that make use of only a small part of the labeled rejects (around 3% for the BS and 9% for the other evaluation metrics). The best dominated BASL variants lying outside of the Pareto frontiers further reduce the number of labeled rejects to between 1% and 3%. This harms the performance compared to the best solutions on the frontiers but still allows outperforming the considered data augmentation benchmarks.

Overall, the results indicate that there is a trade-off between reducing sampling bias and improving score-card performance. This trade-off depends on the quality of the labels assigned to the rejected applications. Naturally, correctly labeling all rejects and appending them to the training data would maximize both the performance and the distribution similarly. In practice, predicted labels of rejects are noisy, which makes labeling too many rejects harm scorecard performance. At the same time, labeling too few rejects does not allow to fully realize the potential of reject inference, as demonstrated by the performance of the dominated BASL scorecards. This bias-accuracy relationship forces a decision-maker to settle for a trade-off. In our paper, we focus on the model accuracy as the ultimate goal of bias correction and tune the meta-parameters of BASL to optimize the scorecard performance.

## Appendix D: Extended Results on Real Data

This appendix provides a more detailed description of the real-world credit scoring data set used in the paper, as well as additional empirical results obtained on that data. First, we demonstrate the presence of sampling bias in the sample of accepted applications and illustrate its impact on scorecard behavior, training, and evaluation. Second, we provide extended results of Experiment I and II. Third, we evaluate the impact of our proposition in settings with the different number of features, simulating the transition between traditional banking and FinTech market.

### D.1. Data Description

The real-world credit scoring data set is provided by a FinTech company Monedo. The data set constitutes consumer microloans issued to customers in Spain between 2013 and 2019. The data contains 59,593 loan applications that split into three groups: (i) 39,579 applications that were accepted by Monedo and granted a loan after scoring them with an existing scorecard; (ii) 18,047 applications that were rejected by Monedo and denied credit after scoring; (iii) 1,967 applications that were accepted by Monedo without scoring.

The set of applicants who have been accepted after scoring serves as historical labeled data that suffers from sampling bias. For all accepts, the data contains the target variable – a binary indicator of whether the customer has timely repaid the loan (*good*) or experienced delinquency of at least three consecutive months (*bad*). The target is not available for rejects, as their repayment outcome is unknown.

The sample with 1,967 customers granted credit without scoring was collected by Monedo to construct an unbiased holdout set. The sample includes applications that would normally be rejected by a scorecard and represents the through-the-door population of customers who apply for a loan. This unbiased sample allows us to evaluate the performance gains from our propositions under the true operating conditions of Monedo.

Apart from the repayment behavior, the data provides 2,409 features used by scorecards. The features split into nine categories summarized in Table D.5. Loan parameters, applicant's attributes and financial data represent information extracted from the application form. Traditional banks frequently use a selected number of features from these categories for credit scoring. Expanding beyond the traditional features, the data contains additional information about loan applicants, including credit bureau data, microgeographic data, data on previous applications submitted by the customer at Monedo and transactional bank data describing the applicant's financial behavior. Finally, the data contains two categories of features describing the device used to submit the loan application and applicant's behavior during the corresponding web session.

| Feature category | No. features | Feature examples |
|---|---|---|
| Loan parameters | 27 | Loan amount, loan duration, etc. |
| Applicant's attributes | 201 | Employment status, number of kids, etc. |
| Applicant's financial data | 8 | Stated income, stated income per household member, etc. |
| Credit bureau data | 60 | Applicant's credit rating, etc. |
| Microgeographic data | 396 | Location of residence, residence type, etc. |
| Previous loan applications | 95 | Number of previous loans, time since the last application, etc. |
| Transactional bank data | 1330 | Average monthly income, average monthly spendings, etc. |
| Internet device data | 210 | Device use to submit application, screen resolution, etc. |
| Online behavior data | 82 | Time to fill out the application form, number of clicks, etc. |

### D.2.    Sampling Bias Illustration

Due to the high dimensionality of the real-world data, we focus on a subset of important features to illustrate sampling bias. First, we remove features that exhibit high multicollinearity by applying correlation-based filtering (Spearman or Pearson correlation above .95), which reduces the number of features from 2,409 to 1,549. Second, we train an XGB-based scorecard to produce estimates of the feature permutation importance. We rank features by their importance and use the most important features in the following analysis.

Figure D.3 illustrates sampling bias and its adverse effects on credit scorecards. Panel (a) compares the distribution densities of the most important feature denoted as $x_1$ on $D^a$, $D^r$ and $H$. The results show clear differences in the feature distribution. This observation is supported by the results of three statistical tests over the top-ten most important features at the 1% significance level. Little's mean-based test rejects the null hypothesis that the labels are missing completely at random, indicating the presence of sampling bias (Little 1988). The Probit-based Lagrange Multiplier test reaches the same conclusion and rejects the null hypothesis of the absence of non-random sample selection (Marra et al. 2017). Finally, the kernel MMD test indicates the difference in the feature distribution between the accepts and the holdout sample (Gretton et al. 2012). Overall, the results indicate that the data exhibits sampling bias in previously accepted applications.

Bias in the training data affects the scorecard behavior. Panel (b) compares the coefficients of the top-five most important features of two exemplary scorecards: (i) biased model $f_a(X)$ trained over $D^a$; (ii) oracle model $f_o(X)$ trained over $H$. Both scorecards use LR as a base classifier. The results indicate that sampling bias in accepts affects coefficients of the trained scorecard, causing them to diverge from the oracle values. The differences are observed in coefficient sizes (e.g., for $x_1$) as well as in their signs (e.g., for $x_4$). The bias in the model parameters translates into a difference in the scores predicted by the scorecards illustrated in panel (c). The accepts-based model $f_a$ provides more optimistic scores compared to $f_o$.

Panel (d) depicts the impact of sampling bias on the scorecard evaluation in four performance metrics. It compares the actual AUC, BS, PAUC and ABR of $f_a$ on the representative holdout sample $H$ (labeled as oracle performance) and the estimated performance of $f_a$ obtained when evaluating it on a subset of accepts. Performance values are aggregated using leave-one-out validation on the holdout sample. The results suggest that evaluating $f_a$ on accepts provides a misleading estimate in all four performance metrics. This implies that the scorecard evaluation on real data is affected by sampling bias and requires correction. Similarly,

**Figure D.3    Sampling Bias Illustration on Real Data**

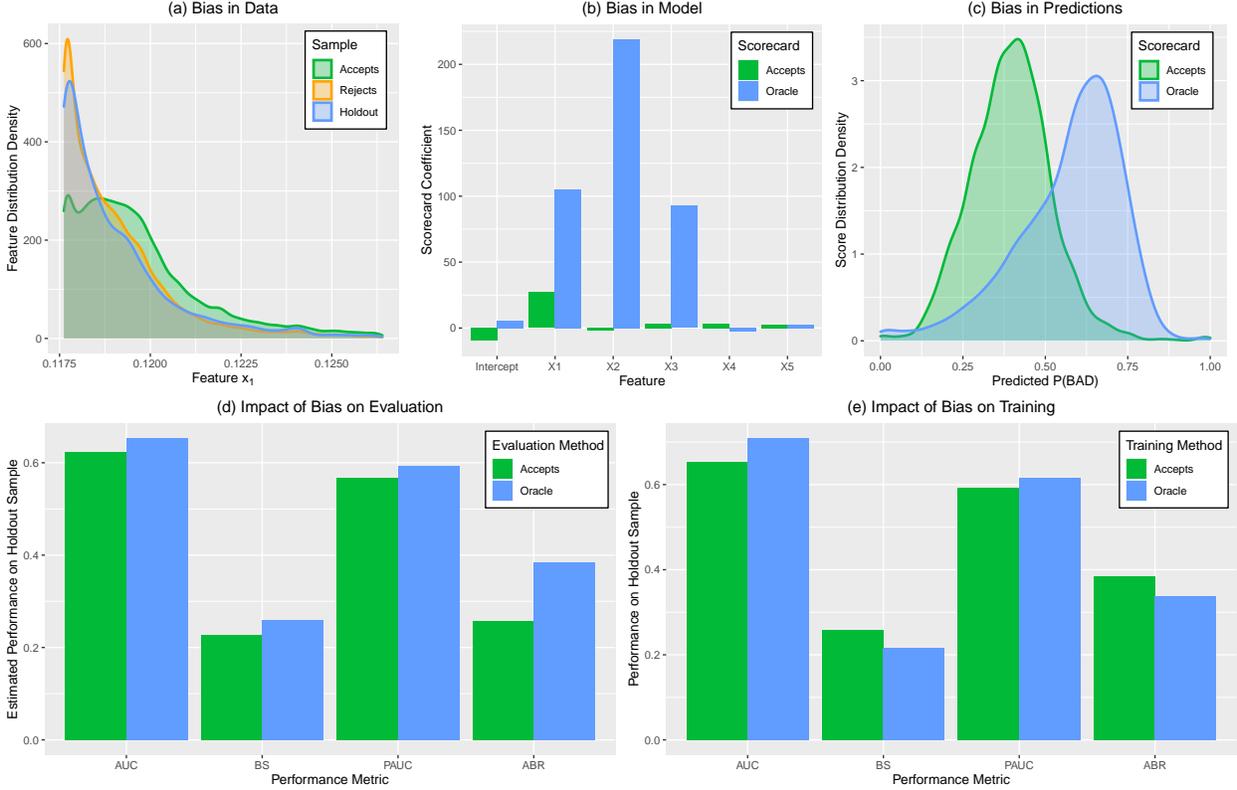

*Note.* The figure illustrates sampling bias and its adverse effect on the scorecard behavior, training and evaluation on real credit scoring data set. Panel (a) demonstrates the bias in the distribution of one of the features. Panel (b) visualizes the bias in scorecards by comparing their coefficients for the top-five features. Panel (c) shows the bias in scorecard predictions for new loan applications. Panels (d) and (e) depict the impact of bias on scorecard training and evaluation in four performance metrics: the AUC, BS, PAUC and ABR.

panel (e) illustrates the adverse effect of sampling bias on the scorecard training. It compares the AUC, BS, PAUC and ABR of $f_a$ and $f_o$. In each of the four metrics, the performance of the scorecard deteriorates when training it on a biased set of accepted applications, indicating that the bias also affects model training.

Overall, the results on real data demonstrated in Figure D.3 are in line with the results on synthetic data presented in Figure 2 in Section 7. The sampling bias originates from the fact that the training data of previously accepted applicants is not representative of the borrowers' population. The bias in the data translates to the bias in model parameters and predictions and negatively affects the training and evaluation of scorecards. Bias correction methods could help to improve both of these dimensions.

### D.3.   Experiment I

This section provides the results of the statistical tests performed in Experiment I on real data. To check the statistical significance of the results presented in Table 2 in Section 7, we perform a Friedman's non-parametric rank sum test for performance differences (Friedman 1937). The null hypothesis of the test is that all evaluation methods have similar performance. The null hypothesis is rejected for all performance

**Figure D.4    Experiment I: Critical Difference Plots for Nemenyi Tests**

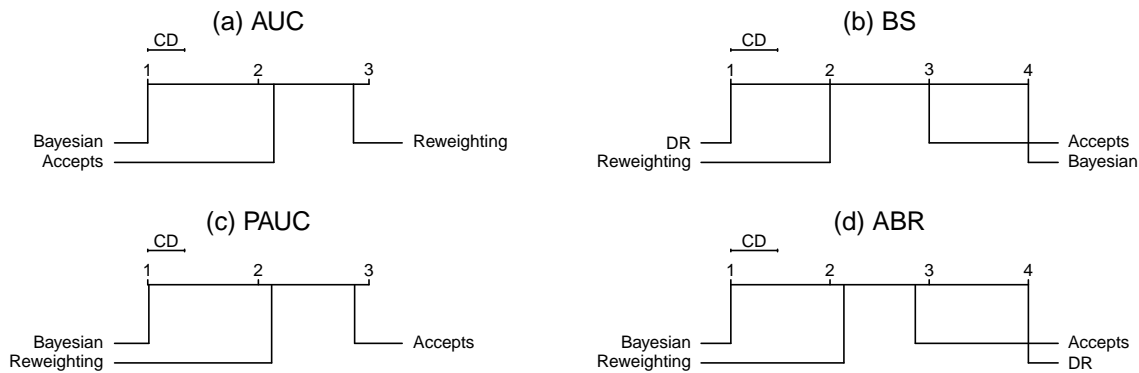

*Note.* The figure depicts rank differences between evaluation methods. The bold segments connect methods for which the differences are not statistically significant at the 5% level according to the Nemenyi post-hoc pairwise test. Abbreviations: AUC = area under the ROC curve, BS = Brier Score, PAUC = partial AUC on FNR $\in [0, .2]$, ABR = average *bad* rate among accepts at 20-40% acceptance.

measures with p-values below $2.2 \times 10^{-16}$. Given that the Friedman test indicates differences in the predictive performance, we proceed with post-hoc tests of pairwise differences between the evaluation methods.

We also use a Nemenyi post-hoc pairwise test, which compares the differences between the average ranks of two methods to the critical difference value determined by the significance level (Demšar 2006). Figure D.4 depicts the rank differences between the evaluation methods based on the Nemenyi test results. The bold segments connect evaluation techniques for which the rank differences in a given evaluation measure are not statistically significant at a 5% level. The results suggest that the Bayesian evaluation framework outperforms both accepts-based evaluation and importance reweighting.

## D.4.   Experiment II

This section provides the results of the statistical significance tests performed in Experiment II om real data and the ablation study that investigates performance gains from different stages of the BASL framework. Similar to Experiment I, we check the significance of the performance gains presented in Table 3 in Section 7. The null hypothesis of the Friedman test that all bias correction methods have similar performance is rejected for all four performance measures with p-values of each test statistic below $2.2 \times 10^{-16}$.

Figure D.5 depicts the rank differences calculated for the pairwise Nemenyi post-hoc tests. As indicated in the figure, BASL outperforms all bias correction benchmarks at a 5% significance level in the AUC, PAUC and ABR. BASL also achieves the best BS, but the BS improvement over the closest competitor, parceling, is not significant at 5% level. In many cases, multiple of the other bias correction benchmarks perform similarly to or worse than ignoring rejects. Parceling and cluster-based reweighting are two methods that tend to come closer to BASL than the other benchmarks in terms of the mean ranks.

Table D.6 provides results of the ablation study of BASL. The table displays incremental performance gains from different algorithm steps, starting from traditional self-learning and incorporating the proposed extensions. The extensions make different contributions to the overall performance of BASL.

**Figure D.5    Experiment II: Critical Difference Plots for Nemenyi Tests**

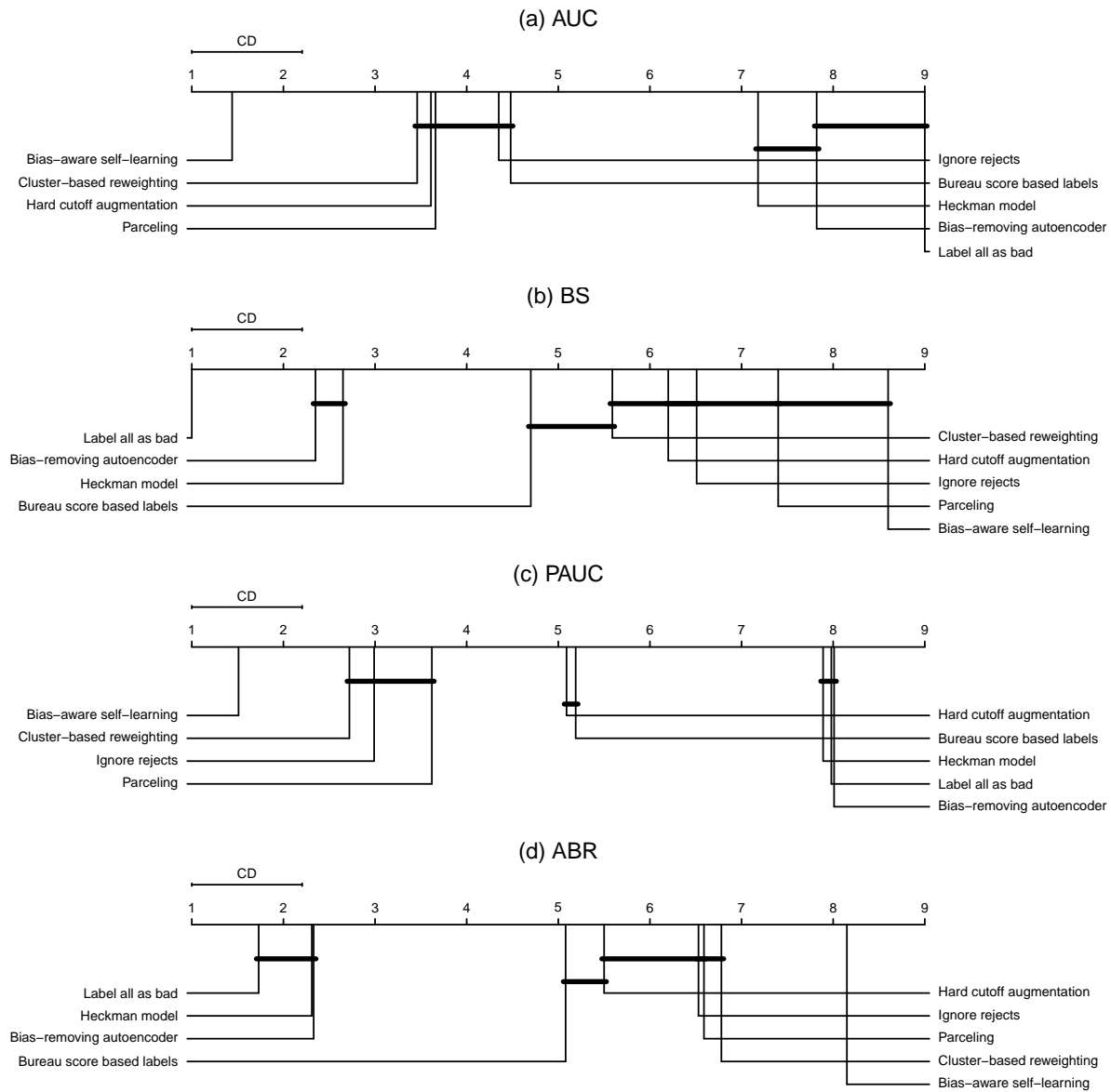

*Note.* The figure depicts rank differences between evaluation methods. The bold segments connect methods for which the differences are not statistically significant at the 5% level according to the pairwise Nemenyi post-hoc test. Abbreviations: AUC = area under the ROC curve, BS = Brier Score, PAUC = partial AUC on FNR $\in [0, .2]$, ABR = average *bad* rate among accepts at 20-40% acceptance.

Overall, incorporating different extensions on top of the traditional self-learning framework improves the the model performance, increasing the PAUC from .6868 to .7075 and the ABR from .2387 to .2211. The largest performance gains in the cost-sensitive metrics are attributed to introducing the filtering stage, which improves the overall rank from 4.80 to 3.93. Gains from implementing the early-stopping mechanism using the

Table D.6     Ablation Study: Gains from Different BASL Steps

| Framework extension | AUC | BS | PAUC | ABR | Rank |
|---|---|---|---|---|---|
| Traditional self-learning | .8059 (.0010) | .1804 (.0004) | .6868 (.0011) | .2387 (.0020) | 4.80 |
| Filter rejects using isolation forest | .8054 (.0011) | .1790 (.0004) | .6981 (.0013) | .2312 (.0022) | 3.93 |
| Label rejects with a weak learner | .8134 (.0006) | .1774 (.0002) | .6992 (.0009) | .2294 (.0011) | 3.60 |
| Introduce the imbalance multiplier | .8133 (.0006) | .1796 (.0002) | .7026 (.0010) | .2238 (.0012) | 3.48 |
| Sampling rejects at each iteration | .8157 (.0006) | .1765 (.0002) | .7035 (.0010) | .2254 (.0013) | 2.85 |
| Bayesian framework for early stopping | **.8166** (.0007) | **.1761** (.0003) | **.7075** (.0011) | **.2211** (.0012) | **2.34** |

Abbreviations: AUC = area under the ROC curve, BS = Brier Score, PAUC = partial AUC on FNR $\in [0, .2]$, ABR = average *bad* rate among accepts at 20-40% acceptance, rank = average rank across the four evaluation measures.

Bayesian evaluation framework are observed in all four evaluation metrics, which emphasizes the important role of using a bias-corrected evaluation metric when performing the model selection.

## D.5. Varying the Number of Features

The real-world credit scoring data set used in this study contains 2,409 features, which is representative of high-dimensional data used for credit scoring by FinTech companies. At the same time, many banks continue using parsimonious scorecards, which comprise a small set of features. To evaluate our propositions across different types of financial institutions, we vary the number of features used by a scoring model and repeat Experiments I and II for different feature groups.

We leverage the feature categories outlined in Table D.5 and iteratively combine features from different groups as depicted in Table D.7. We start by selecting seven basic features with the key information about the loan application, such as applicant income, employment status, and the loan amount. This represents a parsimonious scorecard. Next, we extend the data to 236 features, including more data on the loan parameters, applicant attributes, and financial information. A more sophisticated bank may go beyond these features and purchase third-party data to facilitate more accurate scoring. To reflect such an approach, we iteratively extend the set of features by adding credit bureau data and microgeographic data. Next, we add data on previous loan applications and transactional data describing the applicant's financial behavior. This extends the number of features to 2,117. Finally, simulating a setting where a loan application is submitted online, we add two further groups of features describing the device used by the applicant and the applicant's behavior during the corresponding web session.

Table D.7     Feature Groups

| Group | No. features | Description |
|---|---|---|
| G1 | 7 | Basic features describing the loan applicant |
| G2 | 236 | G1 + loan parameters, applicant's attributes and financial data |
| G3 | 296 | G2 + credit bureau data |
| G4 | 692 | G3 + microgeographic data |
| G5 | 787 | G4 + data on previous loan applications |
| G6 | 2,117 | G5 + transactional bank data |
| G7 | 2,327 | G6 + device data |
| G8 | 2,409 | G7 + online behavior data |

**Figure D.6      Gains from our Propositions Depending on the Data Dimensionality**

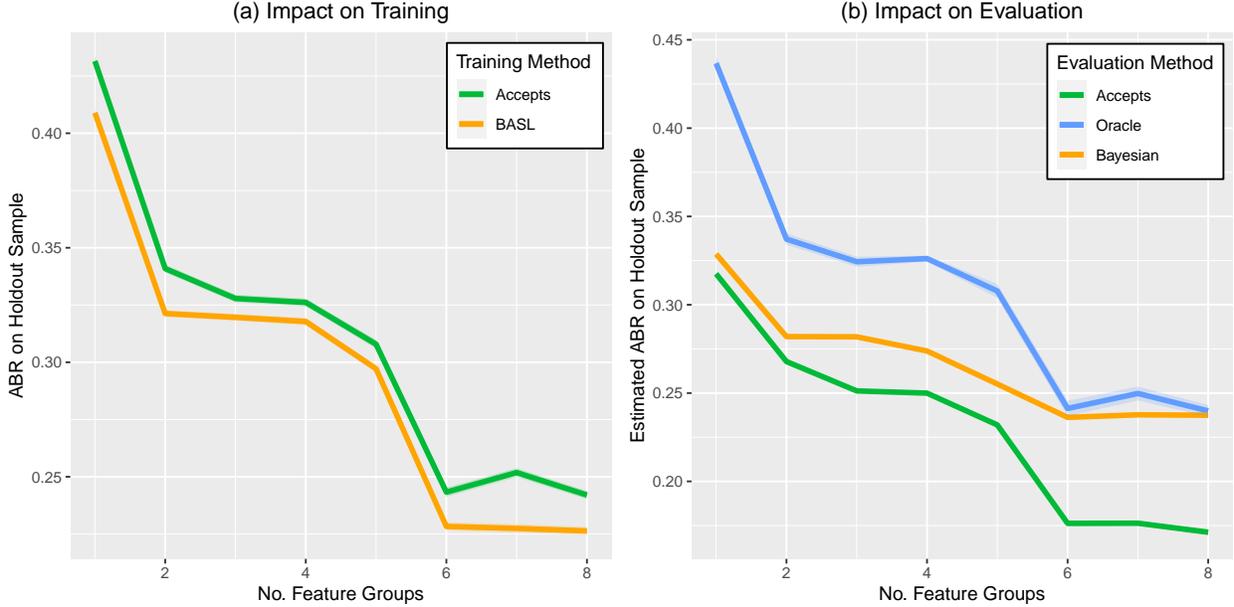

(a) Impact on Training

(b) Impact on Evaluation

*Note.* The figure depicts performance of BASL and Bayesian evaluation depending on the number of feature groups used by a scorecard. Abbreviations: ABR = average *bad* rate among accepts at 20-40% acceptance.

Figure D.6 depicts the results of Experiments I and II in settings with different numbers of features. Panel (a) provides results for BASL against the accepts-based scorecard, which ignores rejects, whereas panel (b) depicts the performance of Bayesian evaluation compared to accepts-based evaluation and the actual performance on the holdout set labeled as oracle. As expected, regardless of the sampling bias correction, the scorecard performance generally tends to improve when more features are available.

Both our propositions consistently improve on ignoring rejects across all settings. At the same time, the largest gains are observed in higher-dimensional data, where the number of features used for credit scoring exceeds 2,000. When using between 6 and 8 feature groups, BASL improves the ABR by up to .0244, whereas the Bayesian evaluation reduces the error in the ABR estimate from .0687 to just .0024. Performance gains on low-dimensional data are more modest. This implies that practitioners in different financial institution types, including FinTechs and traditional banks, may benefit from bias correction, but the larger gains are more likely to be observed in high-dimensional data environments typical for FinTech companies.

## Appendix E:   Meta-Parameters of Data Generation and Bias Correction Methods

This appendix provides meta-parameter values of the base classifiers and bias correction methods. We also provide parameters of the data generation process and the acceptance loop used in the simulation study.

### E.1.   Synthetic Data

The data generation process and the acceptance loop have multiple important meta-parameters. Concerning the data generation, we assume the number of mixture components $C = 2$ and set the distribution parameters as follows: $\boldsymbol{\mu}_1^g = (0,0), \boldsymbol{\mu}_1^b = (2,1), \boldsymbol{\mu}_2^g = \boldsymbol{\mu}_1^g + \vec{1}$ and $\boldsymbol{\mu}_2^b = \boldsymbol{\mu}_1^b + \vec{1}$. The elements of $\boldsymbol{\Sigma}_c^i$ are drawn from a uniform

distribution $\mathcal{U}(0, \sigma_{max})$. We run the acceptance loop for 500 iterations, assuming $n = 100$ and $h = 3,000$. In the MAR setup considered in Section 7.1.1, we set $\alpha = .15$, $b = .70$ and $\sigma_{max} = 1$. In the sensitivity analysis presented in Section 7.1.2, we vary $\alpha$, $\beta$ and $\sigma_{max}$ to investigate the boundary conditions affecting the performance of our propositions. In the MNAR setup considered in Section 7.1.3, we assume $\boldsymbol{\mu}_1^g = (0, 0, 0)$, $\boldsymbol{\mu}_1^b = (2, 1, 0.5)$, $\boldsymbol{\mu}_2^g = \boldsymbol{\mu}_1^g + \vec{1}$ and $\boldsymbol{\mu}_2^b = \boldsymbol{\mu}_1^b + \vec{1}$, hiding the feature with the smallest mean difference from the scorecard and using it for overwriting the scorecard predictions. XGB is used as a base classifier for all scoring models. The meta-parameters of XGB on synthetic data are provided in Table E.8.

Concerning the BASL framework, we set the filtering thresholds $\beta$ to $(.05, 1)$. In the labeling stage, we set $\theta = 2, \rho = .8, \gamma = .01$ and $j_{max} = 3$. We use LR to label the rejected applications and use the Bayesian evaluation framework for early stopping the labeling iterations. To perform the Bayesian evaluation, we set the convergence threshold $\epsilon$ to $10^{-6}$ and specify the number of Monte-Carlo simulations between $10^2$ and $10^4$. The prior on the labels of rejects denoted as $\mathbf{P}(\mathbf{y}^r | \mathbf{X}^r)$ is obtained by predicting the scores of rejected cases using the accepts-based scoring model and calibrating them using LR.

## E.2. Real Data

Table E.8 provides the list of the candidate values and the selected values of the meta-parameters of the XGB classifier that is used as a base classifier for all bias correction methods considered in Experiment I and II on the real data. The meta-parameter values are optimized using grid search on a subset of training data.

Table E.9 contains the bias correction methods considered in the empirical comparison, including both training and evaluation strategies. For each bias correction method, we provide a list of their meta-parameters, including a set of candidate values used for the meta-parameter tuning and the values selected after tuning. Two baseline bias correction strategies – ignoring rejects and labeling all rejects as *bad* risks – do not have any meta-parameters and are not included in the table.

## Appendix F: Implementation of Benchmarks

This appendix provides further implementation details and additional empirical results for some variants of the bias correction benchmarks not included in the paper. The considered benchmarks include importance reweighting techniques, doubly robust evaluation and the bias-removing autoencoder.

**Table E.8    Meta-Parameters of Base Classifiers**

| Meta-parameter | Candidate values | Selected values (synthetic data) | Selected values (real data) |
|---|---|---|---|
| Maximum number of trees | 100, 1,000, 10,000 | 100 | 10,000 |
| Early stopping rounds | 100, no early stopping | no early stopping | 100 |
| Maximum depth | 1, 3, 5 | 3 | 3 |
| Learning rate | .1, .3 | .1 | .1 |
| Bagging ratio | .8, .9, 1 | .8 | .8 |
| Feature ratio | .8, .9, 1 | .8 | .8 |

### F.1. Reweighting

This section focuses on the reweighting techniques considered in this paper. Reweighting tackles sampling bias by estimating importance weights for training examples to rebalance the loss function of the trained algorithm towards examples that are more representative of the population. Given a biased training set $D^a$ and a representative test set $H \subset D$, weights of training examples can be computed as a ratio of two distribution densities: $w(X) = p_H(X)/p_{D^a}(X)$. We focus on the two established families of reweighting techniques: density

**Table E.9    Meta-Parameters of Bias Correction Methods**

| Bias correction method | Meta-parameter | Candidate values | Selected values |
|---|---|---|---|
| Bias-removing autoencoder | Learning rate | .01 | .01 |
| | Number of training epochs | 50, 100 | 100 |
| | Batch size | 50, 100 | 100 |
| | Regularization parameter | $10^{-1}, 10^{-2}, ..., 10^{-5}$ | $10^{-5}$ |
| | Number of hidden layers | 3 | 3 |
| | Bottleneck layer size | $.8k$, $k$ is no. features | $.8k$ |
| Bivariate Heckman | Number of features | 5, 10, ..., 100 | 65 |
| | Functional form | probit, logit, linear | linear |
| Two-Step Heckman | No. features (selection) | 5, 10, ..., 2410 | 30 |
| | No. features (outcome) | 5, 10, ..., 2410 | 2410 |
| | Functional form (selection) | Probit, XGB | Probit |
| | Functional form (outcome) | Probit, XGB | XGB |
| Bureau score based labels | Rating of *good* risks | $\{AA\}, \{AA, A\}$ | $\{AA, A\}$ |
| | Rating of *bad* risks | $\{D\}, \{C, D\}$ | $\{D\}$ |
| Hard cutoff augmentation | Probability threshold | .3, .4, .5 | .5 |
| Reweighting | Truncation parameter | .01, .05 | .05 |
| | Weight scaling | yes, no | yes |
| | Density ratio estimation | cluster-based, LSIF, KLIEP | cluster-based |
| | Number of leaves in DT | 100 | 100 |
| Parceling | Multiplier | 1, 2, 3 | 1 |
| | Number of batches | 10 | 10 |
| Doubly robust evaluation | Truncation parameter | .01, .05 | .05 |
| | Weight scaling | yes, no | yes |
| | Density ratio estimation | cluster-based, LSIF, KLIEP | cluster-based |
| | Number of leaves in DT | 100 | 100 |
| | Reward prediction model | LR, RF | RF |
| | Reward prediction threshold | .1, .2, ..., .9 | .2 |
| Bias-aware self-learning | Filtering thresholds $\beta$ | (0, 1), (.01, .99), (.01, 1) | (.01, 1) |
| | Sampling ratio $\rho$ | 1, .3 | .3 |
| | Labeled percentage $\gamma$ | .01, .02, .03 | .01 |
| | Imbalance parameter $\theta$ | 1, 2, 3 | 2 |
| | Max number of iterations $j_{max}$ | 5 | 5 |
| Bayesian evaluation | Max number of trials $j_{min}$ | $10^2, 10^3, 10^4$ | $10^2$ |
| | Max number of trials $j_{max}$ | $10^6$ | $10^6$ |
| | Convergence threshold $\varepsilon$ | $10^{-6}$ | $10^{-6}$ |
| | Prior calibration | yes, no | yes |

ratio estimation and cluster-based methods. In addition, we propose and use an alternative weight estimation method that uses isolation forest to produce the importance weights.

We implement two prominent density ratio estimation methods: Kullback-Leibler Importance Estimation Procedure (KLIEP, Sugiyama et al. 2007b) and Least Square Importance Fitting (LSIF, Kanamori et al. 2009). These techniques directly estimate the density ratio without explicit estimation of distribution densities $p_H(X)$ and $p_{D^a}(X)$. KLIEP estimates weights by minimizing the Kullback-Leibler divergence between $p_H(X)$ and $w(X)p_{D^a}(X)$. LSIF formulates a least-squares function fitting problem by modeling weights using a linear model: $w(X) = \sum_{l=1}^b \alpha_l \phi_l(X)$, where $\alpha = (\alpha_1, \alpha_2, ..., \alpha_b)$ are parameters to be learned from data, and $\{\phi_l(X)\}_{l=1}^b$ are basis functions such that $\phi_l(X) \leq 0$ for all $X \in D^a$ and for $l = 1, 2, ..., b$.

The cluster-based method estimates weights based on the empirical frequencies of data examples (Cortes et al. 2008). The data spits into $n$ clusters, $C = \{C_i\}_{i=1}^n$. The example weights within each cluster are computed as a ratio of test and training examples in that cluster: $w(C_i) = |C_i \cup D^a| / |C_i \cup H|$. Following the suggestion of Cortes et al. (2008), we use leaves of a fitted decision tree to form the clusters.

Finally, we estimate importance weights using isolation forest, which is a novelty detection algorithm that estimates the normality of each observation by computing the number of splits required to isolate it from the rest of the data (Liu et al. 2008). We fit isolation forest $g(X)$ on the attributes of cases $\mathbf{X}^h$ in the representative sample $H$. Next, we use $g$ to predict similarity scores for the training examples in $D^a$: $\mathbf{s}^a = g(\mathbf{X}^a \subset D^a)$. The predicted similarity scores can be used to judge the likelihood that a certain example comes from the population distribution. The obtained score vector $\mathbf{s}^a$ is then used as importance weights.

In the domain adaptation literature, importance weights are normally computed based on the comparison between a biased training sample and a representative target test sample. In the credit scoring context, we observe two samples: a labeled set of accepted clients $D^a$ and an unlabeled set of rejected clients $D^r$. Both accepts and rejects are biased with respect to the general population of loan applicants $D$. As indicated in Section 6, this paper has access to a representative holdout sample consisting of loans that were granted to a random set of applicants without scoring. This sample can be used as a target sample for estimating the importance weights. However, a representative holdout sample is normally not available as it is very costly to obtain. In this case, a representative validation sample can be constructed by combining the applications that were accepted and rejected by a scoring model during the same time interval.

Before computing the importance weights, we drop features that have Spearman's or Pearson's pairwise correlation higher than .95 to reduce dimensionality and lower the potential noise in weight estimates. The resulting data set contains 1,549 features. Next, we perform 4-fold stratified cross-validation on the data of accepted applicants $D^a$, following the procedure described in Experiment II in Section 6. Within the cross-validation framework, we iteratively estimate the importance weights of the training examples among accepts and rejected examples using one of the considered reweighting techniques.

For each reweighting method, we estimate importance weights in multiple distinct ways. For the density ratio estimation methods KLIEP and LSIF, we estimate density ratios on two sets of samples: (i) comparing the data of accepted applicants and a holdout sample; (ii) comparing the data of accepted applicants and a validation sample constructed from both accepts and rejects. The estimated ratios serve as training weights.

For the cluster-based method, we train two decision trees that split the data into clusters: (i) the first variant is trained over the accepted applicants; (ii) the second variant is trained over the holdout sample. Both decision trees are limited to 100 leaves to ensure that we have enough observations in each cluster. We assign each training example to a cluster in accordance with the leave of the decision tree in which this example falls. Next, we use each of the leaves to compute cluster-specific example weights in two ways: (i) as a ratio between the number of accepted examples and holdout examples in the cluster; (ii) as a ratio between the number of accepted examples and validation examples in the cluster. This gives us four cluster-based reweighting methods employing different clustering models and different weight estimation techniques.

Similarly, we use two variants of the isolation forest: (i) trained over the holdout sample; (ii) trained over the validation sample consisting of both accepts and rejects. Next, we use the trained models to produce similarity scores for the accepted examples. The similarity scores are then used as importance weights.

Before using the estimated importance weights for sampling bias correction, we truncate weights to reduce their variance (Freedman and Berk 2008). The weights are truncated to the interval $[\alpha, \frac{1}{\alpha}]$, where $\alpha$ is tuned using grid search. We also normalize the obtained importance weights using min-max scaling.

The resulting weights are used for both scorecard training and evaluation. First, we use the importance weights for scorecard evaluation within Experiment I. The examples in the validation subset that contains accepts and rejects are used to calculate one of the four performance metrics used in the paper: the AUC, BS, PAUC and ABR. The BS and ABR are reweighted by multiplying the corresponding application-level errors by the importance weights. The weighted AUC is calculated using a technique suggested by (Keilwagen et al. 2014, Hocking 2020). Second, within Experiment II, we use the importance weights for the scorecard development. The importance weights of accepts act as training weights when fitting the XGB-based scorecard on the data from the training folds. The meta-parameters of the base classifiers are provided in Table E.8.

Table F.10 provides the extended results of Experiment II in terms of the predictive performance of a corrected scorecard on the holdout sample. The results suggest that only some of the reweighting techniques improve the scorecard performance compared to a model with no weights. Overall, one of the cluster-based methods outperforms the benchmarks from the density ratio estimation techniques and isolation forest based reweighting. The best performance is achieved when the decision tree that forms the clusters is trained over accepts, whereas the weights are computed as a ratio between the number of accepts and holdout examples in each cluster. Using isolation forest to estimate weights achieves the second-best performance.

The superior performance of the cluster-based reweighting and isolation forest can be explained by the good scalability of tree-based methods in high-dimensional feature spaces. The density ratio estimation methods KLIEP and LSIF produce noisier estimates, which harms the resulting scorecard performance. The cluster-based reweighting demonstrates the best performance when we calculate the importance weights using a time-based validation set constructed of both accepts and rejects. Relying on such a sample is also easier in practice since a representative holdout set is costly to obtain.

## F.2. Doubly Robust

This section provides additional methodological details on the implementation of the doubly robust off-policy evaluation method (DR, Dudík et al. 2014). Due to the differences between the contextual bandit setting

|  |  |  | Table F.10 | Performance of Reweighting Techniques |  |  |
|---|---|---|---|---|---|---|
| Approach | Weights | Sample | AUC | BS | PAUC | ABR |
| No weights | – | – | .7984 (.0010) | .1819 (.0004) | .6919 (.0010) | .2388 (.0019) |
| Density ratio | KLIEP | Holdout | .7930 (.0011) | .1874 (.0005) | .6815 (.0014) | .2543 (.0024) |
|  | KLIEP | Validation | .7950 (.0011) | .1879 (.0005) | .6775 (.0013) | .2570 (.0023) |
|  | LSIF | Holdout | .8033 (.0007) | **.1827** (.0004) | .6916 (.0013) | .2372 (.0021) |
|  | LSIF | Validation | .8027 (.0008) | .1837 (.0004) | .6936 (.0012) | .2365 (.0019) |
| Cluster-based | A/H | Holdout | .7979 (.0010) | .1861 (.0006) | .6895 (.0014) | .2447 (.0024) |
|  | A/V | Holdout | .7988 (.0010) | .1863 (.0004) | .6848 (.0013) | .2473 (.0021) |
|  | A/H | Validation | .8040 (.0008) | .1840 (.0004) | **.6961** (.0012) | **.2346** (.0022) |
|  | A/V | Validation | .7955 (.0012) | .1844 (.0003) | .6884 (.0013) | .2407 (.0022) |
| Isolation forest | SS | Holdout | **.8045** (.0009) | .1837 (.0004) | .6932 (.0013) | .2381 (.0021) |
|  | SS | Validation | .8029 (.0009) | .1845 (.0003) | .6925 (.0013) | .2407 (.0021) |

Weights: A/H = no. accepts divided by no. holdout examples, A/V = no. accepts divided by no. of validation examples, SS = similarity score. Sample: sample used to estimate weights, train isolation forest or clustering algorithm. Performance measures: AUC = area under the ROC curve, BS = Brier Score, PAUC = partial AUC on FNR $\in [0, .2]$, ABR = average *bad* rate among accepts at 20-40% acceptance rate.

considered in the off-policy evaluation literature and the credit scoring setup considered in this paper, using DR for scorecard evaluation requires some adjustments, which we detail below.

In the off-policy evaluation literature, DR is used in a contextual bandit setting. A decision-maker chooses from a set of possible actions and evaluates a policy that determines the assignment of actions. The quality of a policy, or a classifier, is estimated on historical data. In practice, this data is incomplete as every subject has been assigned to exactly one of the possible actions. The reward associated with that action was observed and is available in the data. The (counterfactual) reward corresponding to other actions cannot be observed. To address this, DR combines estimating importance weights, which account for sampling bias in the historical data, with predicting the policy reward for the missing actions. DR produces unbiased estimates if at least one of the two modeled equations is correct (Su et al. 2020).

The off-policy evaluation setting resembles the credit scoring setup to some extent. Rewards in the form of repayment outcomes are observed for accepted applications. The credit scorecard acts as a policy that determines the assignment of actions (i.e., acceptance vs. rejection). However, a substantial difference between the off-policy evaluation setup and credit scoring concerns the availability of information on policy rewards. We can measure the policy reward by the classifier loss (Dudík et al. 2014), which indicates the predictive performance of the scorecard. In the off-policy evaluation setup, a reward from one of the possible actions is available for each subject in the historical data. DR is then used to combine the observed rewards for actions with the observed outcome and predicted rewards for the remaining actions, where the outcomes are missing. In credit scoring, rewards are only observed for applications that have been accepted in the past (i.e., assigned to one specific action). No rewards are observed for applications assigned to other actions (i.e., rejected) as the financial institution never learns the repayment behavior of rejects. This implies that we need to predict the missing rewards for all rejects.

A second limitation of DR in a credit scoring context is associated with the measurement of reward as classifier loss. This measurement implies that the use of DR is feasible only if we can calculate the evaluation

measure on the level of an individual loan. One exemplary loan-level measure is the BS, which assesses a scorecard by calculating the squared difference between the predicted score and a binary label. However, DR is unable to support non-loan level performance measures, including rank-based indicators. Rank-based indicators such as the AUC are widely used in the credit scoring literature (e.g. Lessmann et al. 2015) and regulatory frameworks such as the Basel Capital Accord highlight their suitability to judge the discriminatory power of scoring systems (e.g. Basel Committee on Banking Supervision 2005, Irwin and Irwin 2012). Lacking support for corresponding performance measures constrains the applicability of DR for credit scoring.

In this paper, we implement DR on both synthetic and real-world data. The labeled data of accepted applications is partitioned into training and validation subsets. The training data is used for training the scorecard that is evaluated with DR. The validation subset provides loan applications used for the evaluation. As with the Bayesian evaluation framework, we append rejects to the validation subset to obtain a representative evaluation sample. The repayment outcomes in the validation set are only available for accepts.

As detailed above, DR includes two main components: calculating propensity scores and predicting missing rewards. The first step involves the estimation of propensity scores or importance weights. For this purpose, we use the same method as for the reweighting benchmarks. The comparison of multiple reweighting procedures is described in detail in Appendix F.1. In our experiments, cluster-based weights with weight clipping performs best and is used for the DR estimator. We calculate importance weights for both accepted and rejected applications in the validation subset and store them for the next steps of the DR framework.

The second step involves the calculation of policy rewards, which requires producing a vector of classifier losses for each of the applications in the validation set. To calculate rewards, we score applications in the validation subset with the scorecard evaluated by DR. Next, we compute the rewards for accepts. This procedure depends on the considered evaluation metric. For the BS, the reward is simply the squared difference between the risk score predicted by the scorecard and the actual 0-1 application label. The ABR metric only penalizes the type-II error (i.e., accepting a *bad* applicant). Therefore, for the ABR, we compute the policy reward as a binary variable that equals 1 if the application is predicted to be a *good* but is actually a *bad* risk, and 0 otherwise. Calculating the other two performance measures used in the paper – the AUC and PAUC – is not feasible on the application level, so we only use DR with the BS and the ABR.

Apart from rewards for accepted clients, we also require rewards for rejects. However, since the actual labels of rejects are unknown, we have to predict the rewards for such applications. For this purpose, we train a reward prediction model on the accepted cases from the validation subset and use it to predict rewards for rejects. Reward prediction is performed using a random forest (RF) regressor for the BS metric and using an RF classifier for the ABR. Both models process all applicant features to predict rewards. Since the ABR calculation requires binary rewards, we convert classifier scores into the class predictions using a specified threshold, which we tune to minimize the RMSE of the DR performance estimates.

The final step of the DR framework is calculating the estimate of the scorecard performance based on the computed rewards and importance weights. The actual rewards on accepts and predicted rewards on rejects are multiplied with the weights to correct for the sample selection bias. Next, we aggregate the resulting values across all applications in the validation subset. For the BS, this implies averaging the corrected squared differences over the applications. For the ABR, which has acceptance rate as a meta-parameter, we average the corrected binary error indicators over a certain percentage of applications predicted as least risky.

### F.3. Bias-Removing Autoencoder

In this section, we take a closer look at the performance of the bias-removing autoencoder. The bias-removing autoencoder tackles sampling bias by finding a function that maps features into a new representational space $\mathbf{Z}$ (i.e., $\Phi : \mathbf{X} \rightarrow \mathbf{Z}$) such that distribution of the labeled training data over $\mathbf{Z}$ is less biased and $\Phi(X)$ retains as much information about $X$ as possible. To analyze the performance of this bias correction method in more detail, we compare the predictive performance of four different scoring models that use features coming from the data representations constructed by different autoencoder variants.

The first scoring model $f_a(X)$ serves as a baseline. The model $f_a$ is trained over raw features over a biased sample of previously accepted clients. The next three scorecards are trained over latent features extracted from different autoencoders. The autoencoders are trained using different data samples. First, we train a standard deep stacked autoencoder $a_1(X)$ over $\mathbf{X}^a$ to reconstruct the features of accepted clients. We extract latent features from the bottleneck layer of $a_1$ and use them for training a new scoring model $f_{a_1}(X)$. The scoring model is, therefore, based on the data representation computed on a biased sample of applicants.

Second, we train the autoencoder $a_2(X)$ with the same architecture as $a_1$ but using a training sample constructed of both accepted and rejected applicants $\mathbf{X}^a \cup \mathbf{X}^r$. The scoring model $f_{a_2}(X)$ is trained over the latent features extracted from the bottleneck layer of $a_2$. The extracted features account for the patterns observed on both accepts and rejects, which should improve the performance of the scorecard.

Finally, we train the third autoencoder $a_3(X)$ on $\mathbf{X}^a \cup \mathbf{X}^r$. Compared to $a_2$, $a_3$ includes an additional regularization term that accounts for the distribution mismatch similar to Atan et al. (2018). The regularization term penalizes the mismatch between the distributions of latent features on accepted examples and examples in a validation sample consisting of accepts and rejects from the same time window. This helps the autoencoder to derive latent features that are distributed similarly on the two data samples. After training the autoencoder, we train a scoring model $f_{a_3}(X)$ on the extracted feature representation.

In all three cases, we use a stacked deep autoencoder architecture with three hidden layers. The number of neurons is set to $.9k$ on the first and the last hidden layer and $.8k$ on the bottleneck layer, where $k$ is the number of features in the input data. To facilitate convergence, we preprocess the data before training the autoencoder. First, we drop features that have Spearman or Pearson pairwise correlation higher than .95, reducing the number of features to 1,549. Second, we remove outliers by truncating all features at .01 and .99 distribution percentiles. Third, we normalize feature values to lie within $[0, 1]$ interval. Other meta-parameters of the autoencoder are tuned using grid search; the list of candidate values is given in Table E.9. All scoring models use XGB-based classifier as a base model; the meta-parameters are provided in Table E.8.

As a mismatch penalty, we use the Maximum Mean Discrepancy (MMD, Borgwardt et al. 2006), which measures a distance between distribution means in a kernel space. The MMD is commonly used as a distribution mismatch measure in the domain adaptation literature (e.g. Pan et al. 2011, Long et al. 2014b). The MMD is measured between the latent features of accepts and latent features of the validation sample. The validation sample refers to a time-based representative sample that contains both accepted and rejected clients. The autoencoders only use the features, ignoring the actual labels. Table F.11 reports the results.

Table F.11    Performance of Bias-Removing Autoencoder

| Features | Sample | MMD | AUC | BS | PAUC | ABR |
|----------|--------|-----|-----|----|----|-----|
| Raw | Accepts | - | **.7984** (.0010) | **.1819** (.0004) | **.6919** (.0010) | **.2388** (.0019) |
| Latent | Accepts | - | .7006 (.0034) | .2212 (.0008) | .6066 (.0026) | .3385 (.0023) |
| Latent | Accepts + rejects | - | .7177 (.0020) | .2187 (.0004) | .6170 (.0013) | .3328 (.0024) |
| Latent | Accepts + rejects | + | .7304 (.0011) | .2161 (.0004) | .6376 (.0019) | .3061 (.0036) |

Features: features used to train a scorecard (either raw features or latent features extracted from the bottleneck layer of the autoencoder). Sample: training sample of the autoencoder. MMD: whether the MMD penalty is included in the autoencoder loss function. Performance measures: AUC = area under the ROC curve, BS = Brier Score, PAUC = partial AUC on FNR $\in [0, .2]$, ABR = average *bad* rate among accepts at 20-40% acceptance rate. Standard errors n in parentheses.

First, we compare latent features extracted from $a_1$ trained on accepts and latent features from $a_2$ trained on both client types. The results suggest that the latter set of features leads to a better predictive performance of the eventual scoring model. Furthermore, including the MMD penalty in the autoencoder loss function allows us to extract features that further improve the scorecard's performance. From this comparison, we can conclude that using data of rejected applications and penalizing the distribution discrepancies helps to find a feature representation that suffers less from sampling bias, which has a positive impact on the performance.

At the same time, comparing the performance of the scoring model $f_a$ trained over the original features of accepts to the scoring model $f_{a_1}$ trained over the latent features of the accepts-based autoencoder, we observe a sharp performance drop in all four evaluation measures. This indicates that the predictive power of the latent features constructed by the autoencoder $a_1$ is too low compared to that of the original features. The observed information loss is too large to be offset by the performance improvement from using rejects and adding a distribution mismatch regularizer. This can be explained by a high dimensionality of the feature space, which complicates the reconstruction task.

## F.4.  Heckman Models

In this section, we take a closer look at the performance of multiple variants of the Heckman model. Building on the linear sample selection model proposed by Heckman (1979), the literature has developed many variants of the Heckman-style sample selection model geared towards different settings. Below, we discuss variants of the Heckman model used in this paper, provide additional implementation details and empirical results.

One of the well-known extensions of the classic linear Heckman model is a bivariate probit model proposed by Meng and Schmidt (1985). Their model accounts for the non-random sample selection and is suitable for settings where the outcome variable is binary. The bivariate probit model represents a theoretically sound approach for the credit scoring setup under MNAR and is commonly used in empirical studies on reject inference (e.g., Banasik et al. 2003).

We leverage the bivariate model as one of the Heckman benchmarks used in this study. The selection and outcome equations in the model are estimated jointly using the maximum-likelihood method. To address the challenge of high-dimensional data, we test Heckman model variants with different subsets of features and select the model that demonstrates the best validation performance. For that purpose, we perform feature selection separately for the outcome and selection process. We train two XGB-based scorecards to produce

estimates of the feature permutation importance: (i) outcome model that predicts whether a customer is a *bad* risk; (ii) selection model that predicts whether a customer is accepted. Using both models, we rank features by their permutation importance for the corresponding task. Next, we iterate through the sets of the most important features, using only the top $k$ features to fit the bivariate probit model, varying $k$ in $[5, 10, ..., 100]$. As specified in Table E.9, the best-performing variant of the bivariate probit uses 65 features.

One of the limitations of the bivariate probit model compared to other benchmarks used in this study is reliance on a linear parametric model to describe the outcome and the selection process. To overcome this challenge, we use a two-step estimation procedure that separates the modeling of the two processes into two consecutive steps. First, the selection process is modeled using a probit model. The usage of probit on the first step is required to ensure the normality of the residuals (Heckman 1976). The fitted selection model is used to compute the so-called Mills ratio for each observation, which is calculated as a ratio of the probability density function to the cumulative distribution function for the selection model predictions. On the second step, the inverse Mills ratio is used as an additional feature in the outcome model, controlling for the impact of the non-random sample selection. This estimation procedure allows using a more powerful ML-based algorithm to model the outcome process. At the same time, prior literature (e.g., Galimard et al. 2018) suggests that using the two-stage Heckman may not be theoretically sound due to violation of the residual distribution parameters. Therefore, the two-stage Heckman model faces a trade-off between fulfilling the statistical assumptions and using a more powerful classifier.

This paper uses a two-stage Heckman model as one of benchmarks. Similar to the bivariate probit model, we perform feature selection: in addition to the inverse Mills ration, we use up to 100 most important features in the probit model that is used to model the selection process and select the model achieving the best performance in terms of ABR. The outcome process in the two-step model is modeled using two different classifiers. Apart from a regular probit model traditionally used in such settings, we also employ an XGB classifier with all available features to develop a scorecard with high discriminative power. The XGB classifier uses the same meta-parameters as the ones used by the scoring models resulting from other bias correction benchmarks (refer to Table E.8 for the values).

Finally, we test another variant of the Heckman model that employs an XGB-based classifier with all features to model both selection and outcome equations. Note that replacing a parametric probit classifier on the outcome stage also violates the normality assumption. At the same time, we explore whether using a more powerful ML algorithm to model the selection process may bring an additional performance improvement.

Table F.12 provides the empirical results comparing several variants of the Heckman model. For better visibility, for each of the four model types, we only provide one variant with the selected combination of features. According to the results, only one of the Heckman models marginally improves on the accepts-based scorecard that ignores sampling bias.

Both probit-based models fall behind the accepts-based benchmark. The joint estimation approach demonstrates a slightly better performance, reaching the ABR of .3030. However, replacing the outcome model with the XGB classifier provides a substantial boost to the predictive performance, which allows to reduce the ABR to .2381. Interestingly, using XGB to model the selection process does not provide a further improvement, which can be explained by the violation of the normality assumptions.

Table F.12    Performance of Heckman Models

| Approach | Classifier | NF(S) | NF(O) | AUC | BS | PAUC | ABR |
|----------|-----------|-------|-------|-----|-----|------|-----|
| Ignore bias | XGB | – | 2,409 | .7984 (.0010) | .1819 (.0004) | .6919 (.0010) | .2388 (.0019) |
| Joint model | Bivariate probit | 65 | 65 | .7444 (.0011) | .2124 (.0006) | .6397 (.0010) | .3018 (.0013) |
| Two-step | Probit + Probit | 60 | 61 | .7418 (.0011) | .2167 (.0006) | .6473 (.0010) | .3030 (.0017) |
| Two-step | Probit + XGB | 30 | 2,410 | **.8034** (.0011) | **.1828** (.0004) | **.6934** (.0014) | **.2381** (.0019) |
| Two-step | XGB + XGB | 2,410 | 2,410 | .8027 (.0013) | .1843 (.0004) | .6887 (.0014) | .2473 (.0022) |

Approaches: ignore bias refers to a regular scoring model trained on accepted applicants; joint model is the jointly-estimated Heckman-style bivariate probit model; two-step is a two-step Heckman estimation procedure that uses inverse Mills ratio from the selection model in the outcome model. NF: no. features used to train the selection model (S) or the outcome model (O). XGB = extreme gradient boosting. Performance measures: AUC = area under the ROC curve, BS = Brier Score, PAUC = partial AUC on FNR $\in [0, .2]$, ABR = average *bad* rate among accepts at 20-40% acceptance rate. Standard errors in parentheses.










\end{document}